\documentclass{article}
\usepackage{arxiv}
\usepackage[T2A]{fontenc}
\usepackage{lineno}
\modulolinenumbers[5]

\usepackage{enumerate}
\usepackage{paralist}
\usepackage{amsmath}
\usepackage{url}
\usepackage{makecell}
\usepackage{xcolor}
\usepackage{natbib}
\usepackage[breaklinks=true]{hyperref}
\usepackage{xurl}
\usepackage{comment}
\usepackage{multirow}
\usepackage{bm}
\usepackage{subcaption}
\usepackage{caption}
\usepackage{amssymb}
\usepackage[inline, shortlabels]{enumitem}
\usepackage[russian,english]{babel}
\usepackage{graphicx}
\usepackage{tabularx}
\usepackage{booktabs}

\graphicspath{ {./graphs/} }

\newcommand{\ourmodel}{\textsc{MSDM}}

\newcommand{\task}[1]{the SMM4H 2022 Task 2}

\title{Prompt to Polyp: Medical Text-Conditioned Image Synthesis with Diffusion Models}
\author{
  Mikhail Chaichuk \\
  AIRI, Moscow, Russia \\
  HSE University, Moscow, Russia \\
  \texttt{mvchaychuk@edu.hse.ru} \\
  \And
  Sushant Gautam \\
  Simula Metropolitan Center for Digital Engineering, Oslo, Norway \\
  Oslo Metropolitan University, Oslo, Norway \\
  \texttt{sushant@simula.no} \\
  \And
  Steven Hicks \\
  Simula Metropolitan Center for Digital Engineering, Oslo, Norway \\
  \texttt{steven@simula.no} \\
  \And
  Elena Tutubalina \\
  AIRI, Moscow, Russia \\
  HSE University, Moscow, Russia \\
  Kazan Federal University, Kazan, Russia \\
  \texttt{tutubalina@airi.net} \\ \\
    \texttt{\url{https://github.com/THunderCondOR/ImageCLEFmed-MEDVQA-GI-2024-MMCP-Team}} 
}

\begin{document}
\maketitle

\begin{abstract}
The generation of realistic medical images from text descriptions has significant potential to address data scarcity challenges in healthcare AI while preserving patient privacy. This paper presents a comprehensive study of text-to-image synthesis in the medical domain, comparing two distinct approaches: (1) fine-tuning large pre-trained latent diffusion models and (2) training small, domain-specific models. We introduce a novel model named MSDM, an optimized architecture based on Stable Diffusion that integrates a clinical text encoder, variational autoencoder, and cross-attention mechanisms to better align medical text prompts with generated images. Our study compares two approaches: fine-tuning large pre-trained models (FLUX, Kandinsky) versus training compact domain-specific models (MSDM). Evaluation across colonoscopy (MedVQA-GI) and radiology (ROCOv2) datasets reveals that while large models achieve higher fidelity, our optimized MSDM delivers comparable quality with lower computational costs. Quantitative metrics and qualitative evaluations by medical experts reveal strengths and limitations of each approach.
\end{abstract}


\newpage
\section{Introduction} 

Multimodal medical tasks involve the integration of computer vision (CV) and natural language processing (NLP) to analyze data from images and text \cite{kazerouni2023diffusion,med-tasks-2,med-tasks-3}. A key consideration is ensuring that the representation of images in models accurately corresponds to the provided textual information. Recently, controllable text-to-image generation has emerged as a significant advancement, allowing the creation of medical images that align with specific attributes described in textual prompts, such as anatomical structures and disease types. 

Medical imaging faces persistent challenges, including the scarcity of large annotated datasets due to high costs, laborious annotation processes, and ethical restrictions on the sharing of patient data \cite{chan2020deep,pinaya2022brain,willemink2020preparing}. Furthermore, diverse datasets with a spectrum of anatomical variations and pathological presentations are critical for models that generalize across clinical settings \cite{chan2020deep,armanious2020medgan,kazerouni2023diffusion}. Diffusion models address these issues with their ability to generate realistic synthetic images conditioned on text or metadata, potentially augmenting limited datasets and facilitating privacy-preserving data sharing \cite{chung2022score,karimi2024diffusion}. 

The data demands of modern generative architectures pose significant challenges in medical tasks, where there is a lack of large-scale annotated datasets. Current approaches to address this limitation include (1) adapting pre-trained models from general domains through fine-tuning on specialized medical datasets, and (2) training compact, task-specific diffusion models from initialization. This methodology presents a complex trade-off: while large latent diffusion models (LDMs) like Kandinsky \cite{kandinsky2_main_paper} and FLUX \cite{flux2024} benefit from prior knowledge of natural image distributions, smaller domain-adapted architectures such as text-conditioned Diffusion Probabilistic Models (DDPMs) \cite{diffusionpaper} may better capture nuanced clinical features at reduced computational costs. In this work, we systematically benchmark these strategies, evaluating both resource-intensive fine-tuning of generalized LDMs and the development of lightweight, medically optimized DDPMs. In particular, we introduce MSDM (Medical Synthesis with Diffusion Models), a modified version of DDPM specifically designed for text-to-image synthesis. Based on Stable Diffusion \cite{Rombach}, our approach incorporates three key components: a variational autoencoder (VAE) to compress images into a latent space, a clinical text encoder, and a text-conditioned U-Net \cite{unet-original} that employs cross-attention mechanisms to dynamically align text embeddings with visual features. Our comparative framework examines critical factors including the impact of LoRa (Low-Rank Adaptation) \cite{lora} parameters on training, text and image data augmentation to enhance diversity, and the influence of overall data volume on learning performance.

To evaluate text-to-image synthesis in the medical domain, we leverage two popular datasets that capture distinct clinical and anatomical contexts: (i) Medical Visual Question Answering for the Gastrointestinal Tract (MedVQA-GI) 2024 Challenge Dataset \cite{hicks2023overview}; (ii) Radiology Object in COntext version 2 dataset (ROCOv2) \cite{rocov2}. Hosted as part of the ImageCLEF 2024 benchmarking initiative, the MedVQA-GI data set provides 2000 paired colonoscopy images and 500 written textual prompts authored by clinicians focused on gastrointestinal anatomy and pathology (e.g. ``Generate a picture containing a polyp''). Derived from the Open Access corpus of PubMed Central (PMC), ROCOv2 contains 80000 radiographs and cross-sectional imaging studies paired with radiology-specific captions and medical concepts (e.g., ``CT-scan abdomen showing the bezoar in the pylorus''). Its scale and diversity across imaging modalities (X-ray, MRI, CT) enable a rigorous evaluation of domain generalization in synthesized radiological images. This dual-dataset framework allows us to  assess generative models across two critical clinical domains: colonoscopy imaging, where subtle texture and color variations convey diagnostic significance, and radiology, where structural accuracy and contrast relationships are crucial.

A preliminary version of this work has appeared in~\cite{chaichuk2024mmcp}. In this journal version, we have made several significant improvements, including:
\begin{enumerate}
    \item We incorporate experiments with the state-of-the-art FLUX.1-dev model (12B parameters) in addition to the Kandindsky 2.2 model (2B parameters). Both models were fine-tuned with LoRa.
    \item Beyond initial gastrointestinal-focused experiments, we validate our method on the Radiology Objects in Context v2 (ROCOv2) dataset. This tests generalizability across anatomical regions (e.g., abdominal CT, chest X-ray) and imaging modalities.
    \item An extended review of the literature.
    \item Comprehensive model evaluation using an extended set of metrics, along with human assessment for a more objective analysis. %
    \item A deeper analysis of the applied methods and evaluation criteria, including their comparative analysis and interpretation of results.
    \item  Error analysis of the best-performing model and a discussion of limitations.
\end{enumerate}

All data and code written in support of this publication are publicly available through \texttt{\url{https://github.com/THunderCondOR/ImageCLEFmed-MEDVQA-GI-2024-MMCP-Team}}. See Table \ref{tab:statement} for our statement of significance.

\begin{table}[t!]
\centering
\caption{Statement of Significance}
\label{tab:statement}
\begin{tabular}{p{11cm}}
\hline
\textbf{Problem} \\
\hline
To what extent can text-conditioned diffusion models, pre-trained on natural images, be effectively adapted to synthesize diverse and clinically accurate medical images across different anatomical domains? Does the integration of medical text prompts with imaging data enhance the generalization and fidelity of synthetic outputs in resource-constrained medical settings? \\
\hline
\textbf{What is already known} \\ 
\hline
Medical image synthesis faces challenges due to scarce annotated datasets, ethical constraints, and the need for anatomical diversity. Diffusion models have emerged as promising tools for generating synthetic images conditioned on text or metadata, potentially addressing data scarcity and privacy concerns. Prior work has explored adapting pre-trained latent diffusion models (LDMs) or training compact, domain-specific models, but systematic comparisons of these strategies are lacking. \\
\hline
\textbf{What this paper adds} \\ 
\hline
1) Introduction of MSDM, a medical text-to-image synthesis model based on Stable Diffusion, incorporating a clinical text encoder, VAE, and cross-attention mechanisms to align textual and visual features.
2) Systematic benchmarking of adaptation strategies for generative models, including fine-tuning large LDMs (e.g., FLUX, Kandinsky) with LoRa and training lightweight DDPMs, while evaluating the impact of data volume, augmentation, and computational costs.
3) Validation across two clinical domains (colonoscopy and radiology)
4) Comprehensive evaluation framework with state-of-the-art models, extended metrics, and human assessments by clinicians to rate image plausibility and prompt adherence.\\
\hline
\end{tabular}
\end{table}

\section{Related Work} %

\label{sec:rw}
Generative models, particularly diffusion models, have recently emerged as transformative tools in medical imaging, offering potential solutions for data scarcity, privacy preservation, and training enhancement~\cite{Islam2024Feb,Kazerouni2023Aug,Bengesi2024May}. This section reviews the foundational principles of diffusion models, their application in text-to-image synthesis moving from general domains to specific medical challenges, advancements within key medical imaging subdomains like radiology and endoscopy, techniques for efficient adaptation, and crucial evaluation and ethical considerations.

\subsection{Foundations of Diffusion Models in Image Synthesis}

Diffusion models represent a paradigm shift in generative modeling, demonstrating superior performance over traditional methods like Generative Adversarial Networks (GANs) and Variational Autoencoders (VAEs) in terms of image fidelity and distribution coverage~\cite{Bengesi2024May,Cao2024Feb}. The core concept, inspired by non-equilibrium thermodynamics~\cite{diffusionpaper}, involves a two-stage process: a forward process that gradually adds noise to data, and a learned reverse process that iteratively removes noise starting from a random distribution to generate a sample~\cite{diffusionpaper, pmlr-v139-nichol21a}. Ho et al. (2020) popularized Denoising Diffusion Probabilistic Models (DDPMs) by demonstrating high-quality image synthesis through optimizing a variational bound~\cite{diffusionpaper}.
Key advancements rapidly followed. Improved noise scheduling strategies and the introduction of guidance mechanisms significantly enhanced control and quality~\cite{Hang2024Jul,Choi_2022_CVPR}. Classifier-guidance used an external classifier to steer generation~\cite{Chen2023Jan}, while classifier-free guidance, proposed by Ho \& Salimans (2022), offered more flexibility by allowing conditioning directly within the model, enabling a trade-off between prompt fidelity and realism, which proved crucial for text-to-image tasks~\cite{Ho2022Jul}. A major breakthrough for efficiency and high-resolution synthesis was the development of Latent Diffusion Models (LDMs) by Rombach et al. (2022)~\cite{Rombach}. LDMs operate the diffusion process within a compressed latent space learned by an autoencoder (often a VAE), drastically reducing computational cost~\cite{Preechakul2022}. This architecture, exemplified by Stable Diffusion~\cite{Rombach}, typically incorporates cross-attention mechanisms within its U-Net backbone to integrate conditioning information, such as text embeddings~\cite{AllWorthWords}.

Compared to GANs, diffusion models exhibit greater training stability and significantly better mode coverage, mitigating the common GAN issue of mode collapse where generated samples lack diversity~\cite{Bengesi2024May,Saxena2021May}. Müller-Franzes et al. (2023) provided empirical evidence in the medical domain, showing their Medfusion LDM achieved substantially higher generative recall (diversity) than state-of-the-art GANs across retinal, X-ray, and histopathology datasets, while maintaining comparable or superior precision (fidelity) and generating fewer artifacts~\cite{Muller-Franzes2023Jul}. While VAEs offer stable training, they often produce blurrier outputs compared to diffusion models unless augmented with adversarial components~\cite{DiffuseVAE,Bengesi2024May}. LDMs effectively leverage VAEs for compression but rely on the diffusion process for high-quality reconstruction, overcoming the typical VAE blurriness~\cite{Sadat2024May,He2025Feb}.

\subsection{Text-to-Image Generation: From General to Medical Domains}

Text-to-image synthesis evolved from earlier GAN-based approaches (e.g., StackGAN, AttnGAN) and VQ-VAE combined with transformers (e.g., DALL·E)~\cite{Zhang_2017_ICCV,Xu_2018_CVPR,NEURIPS2019_5f8e2fa1,pmlr-v139-ramesh21a}. However, diffusion models rapidly set new benchmarks. Milestones include OpenAI's GLIDE~\cite{Nichol2021Dec} and DALL·E 2~\cite{Ramesh2022Apr}, Google's Imagen~\cite{Saharia2022May}, and the open-source Stable Diffusion~\cite{Rombach}, Kandinsky~\cite{Razzhigaev2023Oct}, and more recently FLUX~\cite{flux2024}. These models, trained on vast internet-scale image-text datasets, demonstrated remarkable ability to generate diverse, high-fidelity images from complex textual descriptions~\cite{Zhang2024May}.

Applying these powerful general-purpose models to medicine, however, presents unique challenges~\cite{Kazerouni2023Aug}. Studies like Adams et al. (2023) examining DALL-E 2's radiological knowledge found that while basic anatomical structures could be generated, accurately rendering specific pathologies or modalities beyond X-rays was difficult without domain adaptation~\cite{dalle2Radiologoy}. Key challenges include significant domain shift between natural and medical images, the need for high anatomical and pathological accuracy, inherent data scarcity in specialized medical fields~\cite{Altaf2019Jul,Nazir2024Jul,Panayides2020May,Najjar2023Aug,Zhang2024Jan}.

Bridging this gap requires fine-tuning and architectural adaptations. Multimodal encoders like CLIP~\cite{Khan2025Jan} are commonly used for text conditioning, but their general-domain training may not fully capture medical nuances. Research is exploring the use or fine-tuning of domain-specific language models like ClinicalBERT~\cite{Alsentzer2019Apr} or RadBERT~\cite{Yan2022Jun}, and integrating medical ontologies~\cite{Lozano2025Jan} to improve text-image alignment in the medical context. Models like MINIM aim for a unified medical image-text generation capability across modalities~\cite{Wang2025Feb}, demonstrating the potential of large-scale medical fine-tuning.

\subsection{Diffusion Models in Medical Imaging Subdomains}

Diffusion models are being adapted across various medical specialties, with notable progress in radiology and endoscopy.

\subsubsection{Radiology (X-ray, CT, MRI)}
Chest X-ray (CXR) generation from radiology reports is a prominent application. RoentGen~\cite{Chambon2022Nov} demonstrated that fine-tuning Stable Diffusion on CXR-report pairs significantly improves fidelity (FID drop from 47.7 to 3.6) and allows control over described pathologies. Chest-Diffusion~\cite{Huang2024} focused on efficiency, using a domain-specific text encoder and a lighter U-ViT architecture. To enhance precision, approaches like CXRL~\cite{Han2024Mar} integrate reinforcement learning with domain-specific rewards to better align generated images with report details. Diffusion models also extend to 3D imaging. Khader et al. (2023) synthesized realistic 3D MRI and CT volumes, highlighting the importance of slice consistency and demonstrating utility for segmentation task augmentation (boosting Dice score from 0.91 to 0.95)~\cite{Khader2023May}. Text-guided 3D generation (e.g., MedSyn for CT~\cite{Xu2023Oct} and applications in cross-modality synthesis (e.g., MRI-to-CT with MC-IDDPM~\cite{Pan2024Apr}) and image harmonization (DiffusionCT~\cite{Selim2023Jan}) further showcase their versatility.

\subsubsection{Endoscopy and Colonoscopy}
This area, central to our work, leverages diffusion models primarily for data augmentation in polyp detection and classification. Initial work focused on conditioning using segmentation masks to control polyp location and shape (e.g., Macháček et al.~\cite{Machacek2023Jun}, RePoLyp~\cite{Pishva}). A key advancement is text-guided generation. PathoPolyp-Diff~\cite{Sharma2025Feb} uses text prompts to specify polyp pathology (e.g., hyperplastic vs. adenomatous) and imaging modality (e.g., NBI vs. WLI), employing cross-class learning to handle data limitations and demonstrating significant improvements in downstream classification accuracy (+7.9\% balanced accuracy). ArSDM~\cite{Du2023Oct} specifically addressed the challenge of generating small polyps using adaptive loss weighting and semantic refinement, also showing performance gains in detection and segmentation. The ImageCLEFmed 2024 MEDVQA-GI challenge provided a benchmark for text-to-endoscopy generation~\cite{hicks2023overview, Ionescu2024Sep}. Our winning approach in this challenge, fine-tuning models like Kandinsky 2.2 and a custom LDM ("MSDM") using LoRA on the limited MedVQA-GI dataset, validated the feasibility of high-fidelity text-to-image generation for colonoscopy~\cite{chaichuk2024mmcp}. Despite progress, achieving consistent realism in complex mucosal textures and lighting conditions remains an ongoing challenge. Similar techniques are also being applied in laparoscopy, translating simulated images to realistic ones~\cite{Allmendinger2024Aug}.

\subsection{Parameter-Efficient Fine-Tuning (PEFT) Strategies}

Given the large size of foundation models and the often limited availability of medical data, PEFT techniques are crucial for adapting diffusion models effectively and efficiently~\cite{peft, Han2024Mar}. These methods update only a small subset of parameters, reducing computational cost and mitigating overfitting. Low-Rank Adaptation (LoRA)~\cite{Hu2021Jun} is widely used, freezing base model weights and training low-rank update matrices. It proved vital in the MedVQA-GI challenge for adapting Kandinsky~\cite{chaichuk2024mmcp}. Extensions like SeLoRA~\cite{Mao2024Aug} dynamically adjust the rank during training for better adaptation. Textual Inversion~\cite{Gal2022Aug} offers an alternative by learning new "word" embeddings for concepts using very few examples (e.g., <diseaseX>), without modifying model weights, enabling "medical diffusion on a budget"~\cite{deWilde2023Mar}. Other techniques include adding adapter modules or using few-shot methods like DreamBooth~\cite{Ruiz2022Aug}, sometimes in hybrid combinations with LoRA~\cite{Zhang2025Jan}. PEFT methods are "game-changers," enabling specialization of large models on niche medical tasks with modest resources~\cite{Christophe2024Apr}.

\subsection{Evaluation Metrics and Ethical Considerations}

Evaluating synthetic medical images requires a multi-faceted approach beyond visual appeal.
Fréchet Inception Distance (FID) is standard~\cite{Heusel2017Jun}, with low scores (e.g., $\approx$3.6 in RoentGen~\cite{Chambon2022Nov}) indicating high fidelity. However, ImageNet-based FID may miss medical nuances~\cite{Woodland2023Nov}, leading to domain-specific variants (e.g., using medical network features~\cite{Cepa2024Sep}). Generative Precision and Recall assess fidelity and diversity, where diffusion models often show superior recall over GANs~\cite{Cao2024Feb}. Downstream task performance (e.g., improved segmentation Dice scores~\cite{Sudre2017Jul} or classification accuracy provides crucial validation of utility~\cite{Zhang2023Oct}.

\section{Datasets}

\label{sec:data}
\subsection{ImageCLEFmed MEDVQA-GI: gastrointestinal images}%

The primary dataset for model development was the MedVQA-GI development dataset from the ImageCLEFmed 2024 MedVQA-GI task, comprising 20,000 image–text pairs provided by the challenge organizers. Each pair included a gastrointestinal (GI) endoscopy image and a textual prompt describing findings (e.g., polyps), anatomical structures, or procedures. The images were sourced from two public datasets: HyperKvasir \cite{Borgli2020} and Kvasir-Instrument \cite{jha2021kvasir}.

The prompts covered a wide range of medical scenarios, including descriptions such as \textit{``Generate an image containing a polyp''}, \textit{``Generate an image from a gastroscopy procedure''}, and \textit{``Generate an image containing biopsy forceps''}. It is important to note that dataset did not consist entirely of unique entries. It contained 2,000 distinct images and 483 unique text queries, with each image paired with multiple prompts. On average, each image had approximately 10 different suitable descriptions (ranging from 7 to 14), creating diverse combinations. The dataset's relatively small size and specialized domain presented significant challenges for training text-to-image generative models to produce both realistic and varied outputs.

For training and validation, a random 10\% of the images (approximately 200 samples) were selected and removed to form a validation set, along with their associated prompts. This approach ensured a clean separation between training and validation data, minimizing the risk of data leakage.

In our experiments, we also utilized a private dataset hosted at \textit{SimulaMet} to compute the Fréchet Inception Distance (FID). This dataset was employed to benchmark the basic FID performance during experiments through the automated evaluation system. This dataset consisted of two subsets:

\begin{itemize}
    \item \textbf{Single-polyp images}: approximately 2,000 images;
    \item \textbf{Multiple-polyp images}: approximately 8,000 images.
\end{itemize}

For FID evaluation, these subsets were combined and used allowing us to assess the generative model’s performance over broader detection scenarios. 

To ensure consistency with future benchmark standards, the newly released public test set for the ImageCLEFmed 2025 challenge, \textit{Kvasir-VQA-test}~\cite{Gautam2024Oct}, was used for quantitative metrics evaluation (the ones in Table \ref{tab:Automated-Evaluation-Metrics}. This test set\footnote{The set is available at \url{https://huggingface.co/datasets/SimulaMet/Kvasir-VQA-test}}, provides an unseen collection of gastrointestinal endoscopy images with corresponding prompts.

The combination of these datasets enabled a thorough evaluation of model performance across varying dataset sizes and complexities, providing insights into the scalability and generalization capabilities of the developed generative models.

\subsection{ROCOv2} %

To evaluate the generalization capability of our approach across medical domains, we conducted additional experiments using the ROCOv2 dataset, a well-established benchmark containing 80,000 image-text pairs of X-ray images with corresponding textual descriptions. The dataset is divided into three subsets: a training set of 60,000 pairs, a validation set of 10,000 pairs, and a test set of 10,000 pairs. Compared to the MedVQA-GI dataset, ROCOv2 provides significant advantages for model evaluation, including substantially larger scale, greater diversity of medical imaging content, and broader coverage of anatomical regions beyond gastrointestinal imaging. This expanded scope enables more robust assessment of our models' ability to generalize across different clinical contexts and imaging modalities.

This evaluation allows us to thoroughly assess our models' performance when trained on more comprehensive medical data while testing their generalization to non-GI medical images.

\section{Methodology} %

\label{sec:methodology}
\begin{figure}[t!]
    \centering
    \includegraphics[width=1\linewidth]{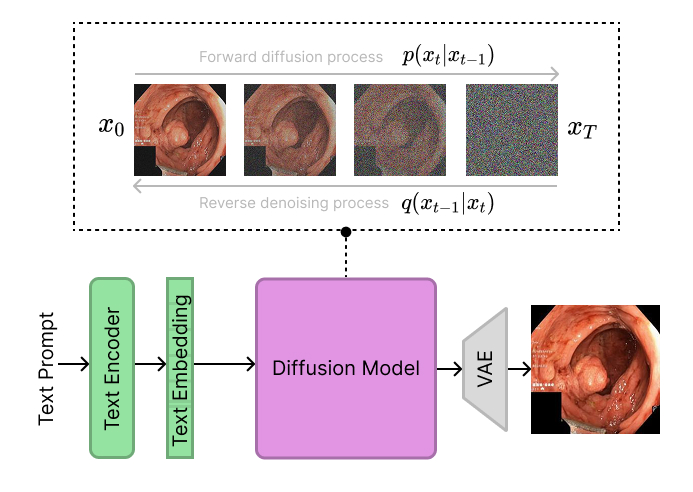}
    \caption{A simplified overview of the text-to-image diffusion model pipeline, commonly used in most modern architectures.}
    \label{fig:diffusion_pipeline}
\end{figure}

In this study, we focused on exploring the capabilities of diffusion models based on Stable Diffusion and U-Net architectures, as they represent the current standard in generative modeling and demonstrate the most promising results. These architectures underlie the majority of modern research and popular open-source solutions, confirming their versatility and efficiency. Figure \ref{fig:diffusion_pipeline} shows the overall architecture of text-to-image diffusion models.

Today, there are numerous large-scale generative models built on diffusion and U-Net principles that achieve high-quality generation across various tasks. Thanks to training on extensive and diverse datasets, such models possess a deep "understanding" of complex structures and patterns. However, in highly specialized domains like medicine, where precision and detail are critical, models fine-tuned for specific subtasks are more commonly used.

In particular, medical imaging (X-rays, MRI, CT scans, etc.) traditionally relies on separate models for each modality, complicating the development process and requiring meticulous data preparation and architecture optimization.

We aimed to investigate whether modern generative models, with their significant out-of-the-box capabilities, can effectively address specialized tasks without resource-intensive training from scratch. Specifically, we examined whether high-quality and diverse generation could be achieved using Parameter-Efficient Fine-Tuning (PEFT) methods, which minimize computational costs.

To test this, we evaluated models of different scales, allowing us to assess their efficiency under limited-resource conditions. The study included:

\begin{itemize}
    \item Kandinsky 2.2 \cite{kandinsky2_main_paper} (2B parameters) – a compact model suitable for tasks with moderate computational requirements.
    \item FLUX.1-dev \cite{flux2024} (12B parameters) – a larger architecture to evaluate the potential of big models in specialized domains.
\end{itemize}

Furthermore, to establish a comprehensive performance baseline contrasting with parameter-efficient fine-tuning methods, we introduced MSDM -- a compact model (500M parameters) that was trained from scratch and has an architecture derived from a proven design that has shown strong performance in medical imaging tasks.

\subsection{Large Pretrained Diffusion Models}
\subsubsection{Kandinsky 2.2}%

The Kandinsky 2.2 (Kandinsky) model was selected for this study due to its favorable balance between generation quality and computational efficiency. It demonstrates competitive performance \cite{kandinsky-top}, despite its relatively small size and simple architecture. This results in significantly lower computational costs and faster training times compared to larger alternatives. For medical applications where both resource efficiency and reliable performance are crucial, these characteristics make Kandinsky 2.2 particularly suitable. We wanted to test whether this compact model could handle medical data generation tasks and compare its performance against much larger models while keeping its computational benefits.

\begin{figure}[t!]
    \centering
    \includegraphics[width=1\linewidth]{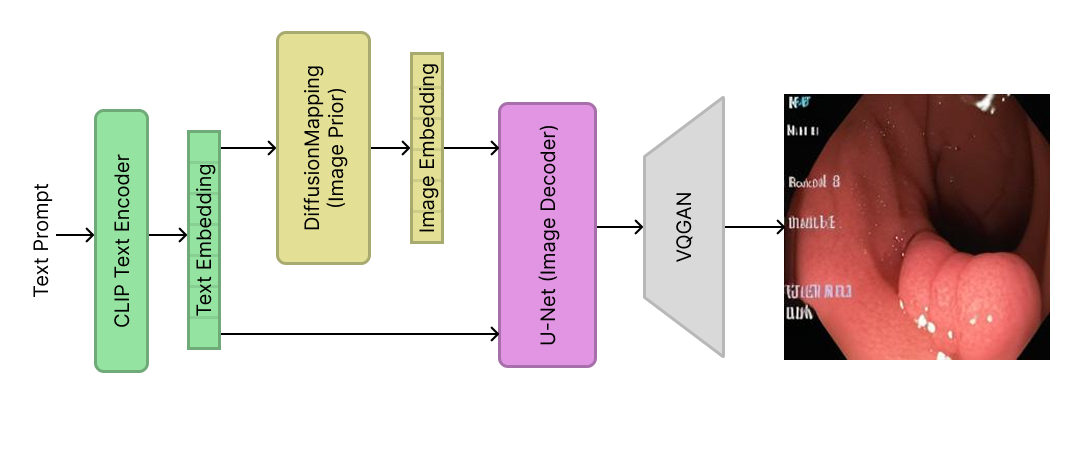}
    \caption{Overview of Kandinsky 2.2 architecture.}
    \label{fig:kandinsky_architecture}
\end{figure}

The architecture of Kandinsky 2.2 differs from the standard pipeline presented above. Its overall structure is shown in Figure \ref{fig:kandinsky_architecture} and consists of the following components:

\begin{itemize}
    \item[1.] \textbf{Image Prior Model}: The DiffusionMapping model \cite{diffusion-mapping} transforms a given text prompt (or its CLIP \cite{clip} text embedding) into a visual embedding within a latent space. The CLIP model used is CLIP-ViT-G.
    \item[2.] \textbf{Image Decoder}: A U-Net-based diffusion model \cite{unet-original} reconstructs images from the latent representations produced by the Image Prior. This stage is trained to perform reverse diffusion, generating images conditioned on the visual embeddings.
    \item[3.] \textbf{Sber-MoVQGAN} The final component is a modified version of VQGAN \cite{vqgan}, optimized by the authors \cite{sber-vqgan} for enhanced reconstruction quality. 
\end{itemize}

The Image Prior model converts text embeddings into visual embeddings, creating a link between textual and visual representations. This dual pipeline allows the decoder to process both text and image data during training. By using both types of input simultaneously, the model improves the overall generation quality through richer semantic features in the latent space.

We used the official model checkpoints \cite{kandinsky-2-prior-weights, kandinsky-2-decoder-weights} hosted on Hugging Face \cite{hugging-face} during all experiments.

\subsubsection{FLUX.1-dev}%

FLUX.1-dev (FLUX) is one of the most advanced open-source text-to-image models, characterized by its impressive scale (12B parameters) and state-of-the-art architectural design. This combination results in superior generation quality, making FLUX.1-dev a popular and widely researched solution in the field.
Given FLUX.1-dev's leading position in image synthesis, we aimed to verify two key aspects: whether its generation quality shows measurable superiority over competing approaches in our specific task, and how its LoRA fine-tuning behavior compares to smaller architectures regarding training stability, hyperparameter sensitivity, and output consistency across generations.
For fine-tuning FLUX, we also utilized weights available on HuggingFace \cite{flux-weights}.

\subsection{Fine-tuned Diffusion Models}

\subsubsection{MSDM}%

In addition to enabling efficient training of large-scale models, we proposed a smaller yet effective denoising diffusion probabilistic model (DDPM) for text-conditioned medical image generation. Our model, \ourmodel, builds upon Medfusion \cite{medfusion} -- an architecture derived from Stable Diffusion \cite{Rombach} -- and consists of two core components: a variational autoencoder (VAE) \cite{vae} for latent space compression and a U-Net \cite{unet-original} for iterative denoising.

Unlike the original Medfusion framework, which was designed for class-conditional generation, \ourmodel incorporates text guidance through the following modifications:
\begin{itemize}
    \item Cross-Attention layers integrated into the U-Net blocks, enabling dynamic interaction between text embeddings and visual features during denoising.
    \item A FeedForward layer after each CrossAttention layer to stabilize gradient flow and enhance feature transformation.
\end{itemize}

For text encoding, we employed the CLIP ViT-L/14 model \cite{clip-site}, which has become the de facto standard in diffusion frameworks due to its balanced performance profile. CLIP offers particular advantages for medical image generation: its embeddings not only encode textual information but also maintain strong alignment with the visual domain. While larger language models exist, CLIP proved particularly suitable for our task as the structured nature of medical captions in our dataset did not require complex linguistic processing, and its joint image-text embedding space inherently supports cross-modal interactions critical for conditional generation.

\subsection{Text Diversity Enhancement with Paraphrasing}

A lack of text diversity in the MedVQA-GI training dataset could lead to model overfitting, causing the generation of overly similar images for different input texts. To address this issue, we explored an approach to increase text variation using a generative language model.
We used GPT-3.5 Turbo to create paraphrased versions of the original texts, expanding the set of unique descriptions from 483 to over 11,000. This approach introduced greater linguistic variability while preserving the semantic meaning of the original data.

To assess the effect of paraphrasing, we tested three strategies:

\begin{itemize}
    \item Simply adding paraphrased texts to the original set.
    \item Randomly substituting some original texts with paraphrased versions.
    \item Completely replacing the original texts with paraphrased ones.
\end{itemize}

Our experiments showed that retaining the original texts while adding paraphrased variants achieved the best results. The other two strategies, particularly full replacement, led to a decline in model performance when evaluated on the original texts. This indicated that supplementing rather than replacing the original descriptions maintained a better balance between diversity and accuracy.

\section{Experiments} %

\subsection{Evaluation Setup and Metrics}
\label{sec:eval}

In this work, as in the preliminary study, we used the FID metric as the primary metric for evaluating generation quality. This allows us to compare new results with those obtained earlier and assess the performance of different models during training.
For the ROCOv2 dataset, the FID was calculated across the entire test set.
For the MedVQA-GI dataset, given the strong dependence of FID on sample size, we computed the metric using all 2000 images from the training set during the training process. Additionally, thanks to our collaboration with the organizers of the MedVQA-GI competition, we were able to evaluate the models on a closed test dataset by submitting generated images to their testing system. For this purpose, 5000 images were generated by each model based on test prompts and subsequently submitted for automated evaluation.

However, since FID does not fully capture all aspects of generation and is a rather unstable metric, in this extended study we expanded the set of evaluation metrics and also invited a medical expert for human evaluation. Expert assessment is particularly crucial in the field of medical imaging, where not only visual quality but also clinical relevance must be considered. Such a combined approach provides a more comprehensive understanding of generation quality and its suitability for real-world applications.

\subsubsection{Human Expert Evaluation Protocol}

We designed a structured human evaluation involving a medical expert with extensive experience in colonoscopy. The expert rated image outputs based on clinical relevance, fidelity to prompt content, and visual quality, supported by a dedicated annotation interface.
A total of \textbf{40 prompts} were used: \textbf{20 original prompts} (even indices) and \textbf{20 semantically rephrased prompts} (odd indices).

For each prompt, the evaluation interface first displayed a set of real colonoscopic images as reference from the test set. Following that, the expert was shown the outputs of three anonymized image generation models — Model A, B, and C — with 10 images per model.

\subsubsection{Annotation Process}

Prior to the annotation session, the doctor was thoroughly briefed on the evaluation criteria and protocol using a rubric-based guide. Consistent understanding across prompts was maintained throughout the session.
The expert rated each model’s image set across the following six evaluation aspects, using a 0--10 scale:

\begin{enumerate}
    \item \textbf{Clinical Realism} -- Visual realism as authentic colonoscopy images.
    \item \textbf{Prompt Faithfulness} -- Alignment with semantic content of the prompt (e.g., abnormality type, text presence, tool count).
    \item \textbf{Detectability of Abnormality} -- Clarity of visualized abnormalities (e.g., polyps, esophagitis).
    \item \textbf{Color and Contrast} -- Use of clinically appropriate color cues and contrast.
    \item \textbf{Intra-set Diversity} -- Diversity across generated images for the same prompt, while remaining clinically plausible.
    \item \textbf{Confidence of Use (Clinical)} -- Suitability of the images for use in clinical training or education.

\end{enumerate}

The doctor was also asked to provide a \textbf{Global Preference} — an overall ranking of the three models based on realism and clinical utility. 

\subsubsection{Quantitative Evaluation Metrics}

\begin{table}[htbp]
\centering
\caption{Automated Evaluation Metrics}
\label{tab:Automated-Evaluation-Metrics}
\begin{tabular}{lp{4cm}p{4.5cm}}
\hline
\textbf{Metric} & \textbf{Description} & \textbf{Computation Method} \\
\hline
Fidelity ($\uparrow$) & Realism vs. real colonoscopy images & Inverse FID: \( \frac{1000}{1 + \text{mean-FID}} \), using BiomedCLIP \\
\hline
Agreement ($\uparrow$) & Consistency between original and rephrased prompts & Cosine similarity between BiomedCLIP embeddings \\
\hline
Diversity ($\uparrow$) & Variability among generated images per prompt & Mean pairwise distance between 10 images per prompt \\
\hline
FBD ($\downarrow$) & Overall realism vs. real images across all prompts & Standard FID (global), using BiomedCLIP features \\
\hline
\end{tabular}
\end{table}

In addition to expert ratings, we computed four quantitative metrics per model using \textbf{BiomedCLIP}~\cite{zhang2024biomedclip} image embeddings. These metrics—Fidelity, Agreement, Diversity, and FBD (Frechet BiomedCLIP Distance)—leverage domain-specific embeddings to provide a semantically meaningful assessment of generated outputs. Unlike traditional benchmarks such as FID, which are optimized for natural images and often overlook critical diagnostic features, our metrics are tailored to the unique demands of medical imaging.

Except for FBD, all metrics are computed per prompt, using sets of real and generated images grouped by their corresponding prompts. This allows us to assess prompt-specific realism (Fidelity), consistency across prompt rephrasings (Agreement), and intra-prompt variability (Diversity). FBD, by contrast, is computed globally across all prompts, capturing the overall distributional alignment between the generated and real datasets.

As shown in Table~\ref{tab:Automated-Evaluation-Metrics}, these scores complement expert evaluations and enable robust benchmarking of model performance. By capturing realism, prompt adherence, variability, and distributional alignment in a clinically relevant embedding space, this evaluation framework supports the safe and effective integration of generative models into medical and educational workflows.

\subsection{Experimental Setup}

All models were trained and validated on images of size $256\times 256$, a resolution that optimally balances computational memory constraints and output quality in our case. We used the AdamW optimizer~\cite{adamw} for optimization with default parameters. Through empirical validation, a learning rate (lr) of $10^{-4}$ was consistently optimal across experiments. Additional hyperparameters were tuned where necessary.

For Kandinsky 2.2, we trained the Prior and Decoder models with LoRA independently, then computed metrics on the combined pipeline. To simplify the overall experimental design, we used the same LoRA rank for both Prior and Decoder models during evaluation.

For MSDM, we first trained the VAE using the hyperparameters suggested by the authors of the base paper. Subsequently, using this trained VAE, we proceeded to train the UNet with its default parameters as well.

The experiments were conducted on the HSE University cHARISMa supercomputer cluster \cite{cluster-link}. For all experiments, we used 1 to 4 NVIDIA A100 GPUs.

 \subsection{Results}
\label{sec:res}

\subsubsection{MedVQA-GI results}
\label{sec:results_1}

\begin{figure}[htbp]
\centering
\subcaptionbox{Original images from the Development Dataset}{
\includegraphics[width=0.15\textwidth]{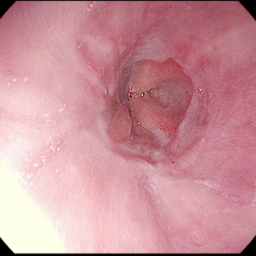} \includegraphics[width=0.15\textwidth]{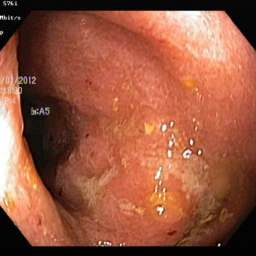} \includegraphics[width=0.15\textwidth]{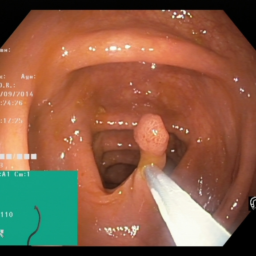} \includegraphics[width=0.15\textwidth]{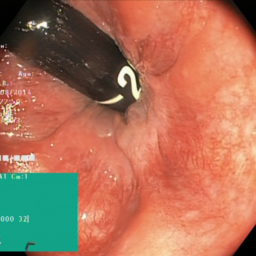} \includegraphics[width=0.15\textwidth]{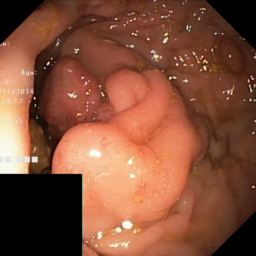}
\includegraphics[width=0.15\textwidth]{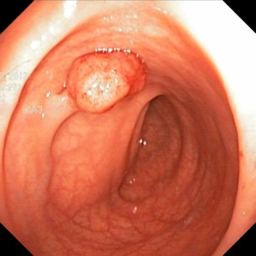}
}%
\hspace{20mm}
\subcaptionbox{FLUX-1.dev}{
\includegraphics[width=0.15\textwidth]{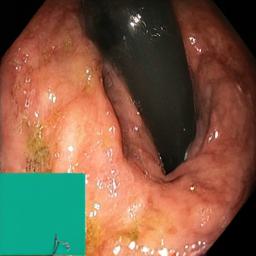} \includegraphics[width=0.15\textwidth]{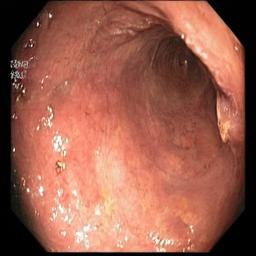} \includegraphics[width=0.15\textwidth]{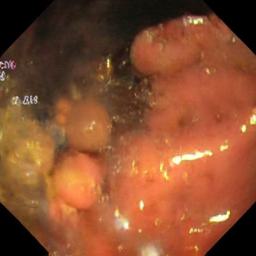} \includegraphics[width=0.15\textwidth]{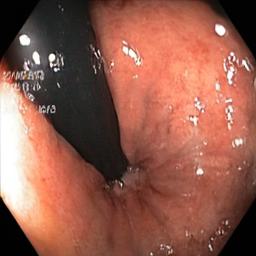} \includegraphics[width=0.15\textwidth]{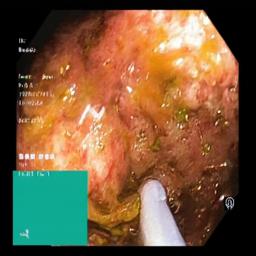}
\includegraphics[width=0.15\textwidth]{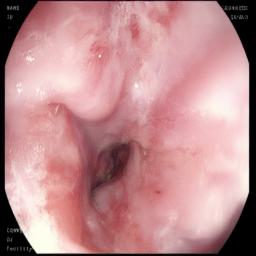}
}%
\hspace{20mm}
\subcaptionbox{Kandinsky 2.2}{
\includegraphics[width=0.15\textwidth]{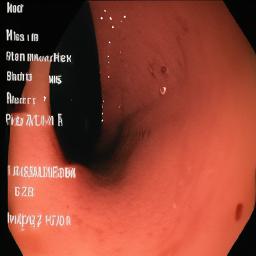} \includegraphics[width=0.15\textwidth]{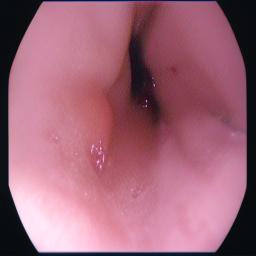} \includegraphics[width=0.15\textwidth]{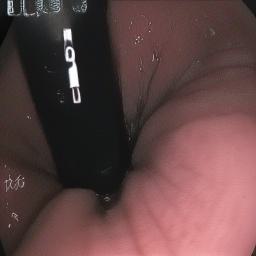} \includegraphics[width=0.15\textwidth]{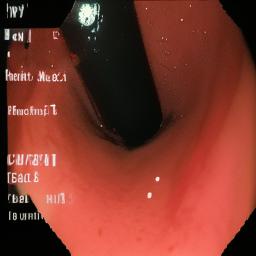} \includegraphics[width=0.15\textwidth]{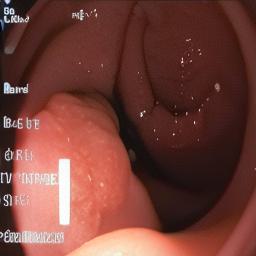}
\includegraphics[width=0.15\textwidth]{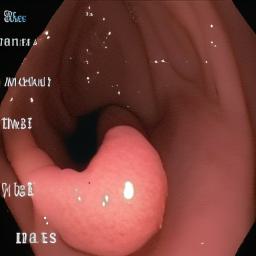}
}%
\hspace{20mm}
\subcaptionbox{MSDM}{
\includegraphics[width=0.15\textwidth]{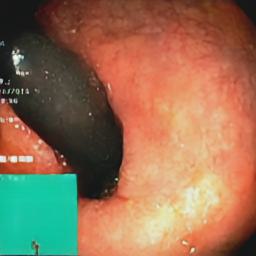} \includegraphics[width=0.15\textwidth]{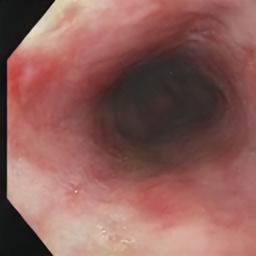} \includegraphics[width=0.15\textwidth]{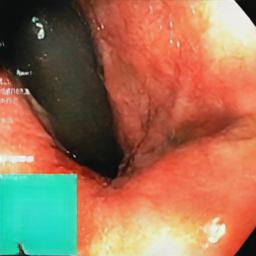} \includegraphics[width=0.15\textwidth]{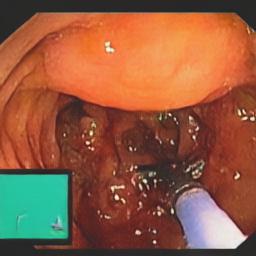} \includegraphics[width=0.15\textwidth]{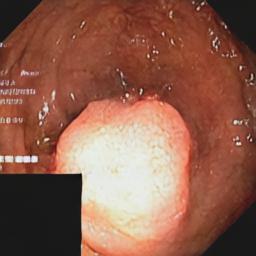}
\includegraphics[width=0.15\textwidth]{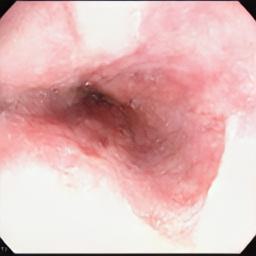}
}%
\caption{Examples of images generated using FLUX, Kandinsky 2.2 and MSDM compared to the original images from the development dataset.}
\label{fig:models_pictures_comparison}
\end{figure}

We conducted systematic experiments to optimize hyperparameters for each model, including dataset augmentation through paraphrasing to enhance both volume and diversity. Based on comprehensive evaluation, we selected the top-performing models that demonstrated optimal generation quality in our assessment. Final model selection was determined by their FID scores on the development dataset. After optimization, we selected LoRa rank 64 for the FLUX-1.dev and Kandinsky 2.2 models.

\begin{table}[ht]
\centering
\caption{Comparison of finetuned FLUX, Kandinsky, and MSDM models across different evaluation metrics.  
Fidelity, Agreement, Diversity, and FBD metrics were computed on the public Kvasir-VQA test set.}
\label{tab:model-comparison}
\begin{tabular}{lccc|cccc}
\toprule
\textbf{Model} & \textbf{FID (Dev)} & \textbf{FID (Test)} & & \textbf{Fidelity} & \textbf{Agreement} & \textbf{Diversity} & \textbf{FBD} \\
\midrule
FLUX       & 45--50     & 72.96 $\pm$ 4.05   & & 0.36 & 0.79 & 0.64 & 1056.81 \\
Kandinsky 2.2  & 100--110   & 121.12 $\pm$ 10.87 & & 0.26 & 0.80 & 0.52 & 2060.58 \\
MSDM       & 50--60     & 94.92 $\pm$ 10.48  & & 0.30 & 0.77 & 0.63 & 1531.62 \\
\bottomrule
\end{tabular}
\end{table}

As shown in Table~\ref{tab:model-comparison}, all models exhibited performance degradation on the private test set compared to the development set. The reported FID scores are averages from five independent runs per model, each based on 5,000 generated images. Notably, dataset paraphrasing during training contributed to a 10–20 point reduction in test FID scores across all models.

Figure \ref{fig:models_pictures_comparison} presents a comparative analysis of the generated outputs from each model alongside the original images from the MedVQA-GI development dataset. Visual evaluation shows that the generated images closely resemble the originals in terms of composition and structure. No significant artifacts or distortions were observed, indicating robust performance across all tested architectures. However, Kandinsky 2.2 may have a slight disadvantage in preserving fine-grained details compared to FLUX-1.dev and MSDM, as its generations are slightly worse in terms of colors and textures, which correlates with the resulting FID values.

\begin{figure}[h!]
\centering
\subcaptionbox{Original images from the ROCOv2 Dataset}{
\includegraphics[width=0.15\textwidth]{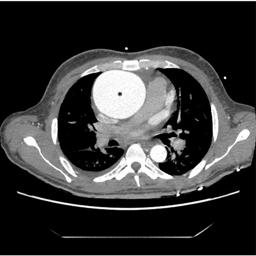} \includegraphics[width=0.15\textwidth]{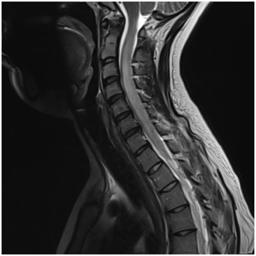} \includegraphics[width=0.15\textwidth]{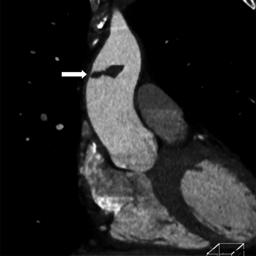} \includegraphics[width=0.15\textwidth]{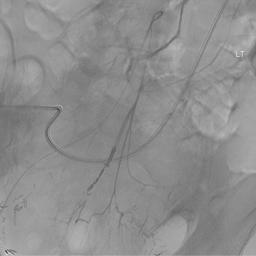} \includegraphics[width=0.15\textwidth]{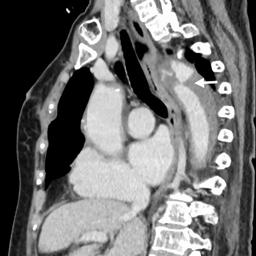}
\includegraphics[width=0.15\textwidth]{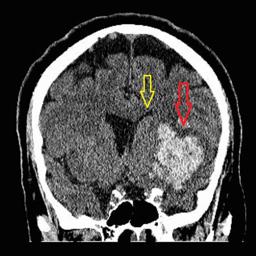}
}%
\hspace{20mm}
\subcaptionbox{FLUX-1.dev}{
\includegraphics[width=0.15\textwidth]{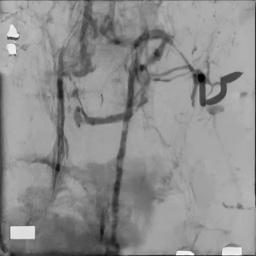} \includegraphics[width=0.15\textwidth]{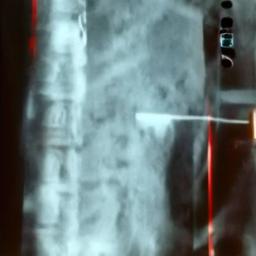} \includegraphics[width=0.15\textwidth]{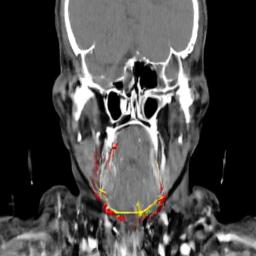} \includegraphics[width=0.15\textwidth]{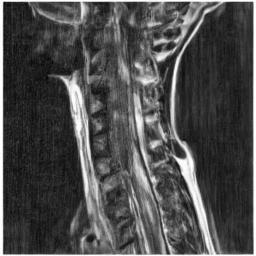} \includegraphics[width=0.15\textwidth]{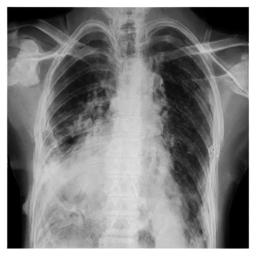}
\includegraphics[width=0.15\textwidth]{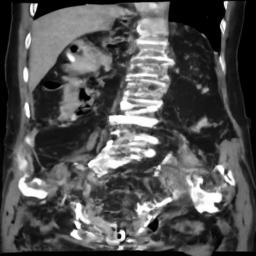}
}%
\hspace{20mm}
\subcaptionbox{Kandinsky 2.2}{
\includegraphics[width=0.15\textwidth]{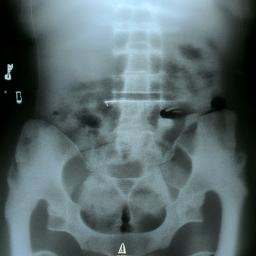} \includegraphics[width=0.15\textwidth]{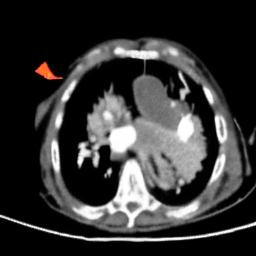} \includegraphics[width=0.15\textwidth]{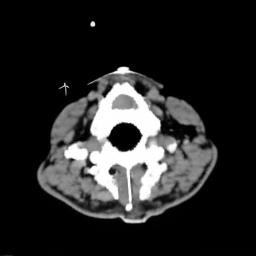} \includegraphics[width=0.15\textwidth]{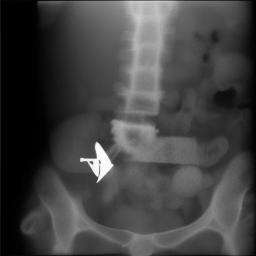} \includegraphics[width=0.15\textwidth]{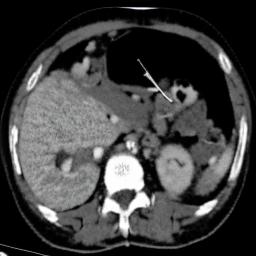}
\includegraphics[width=0.15\textwidth]{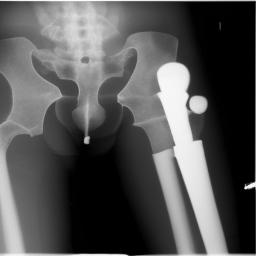}
}%
\hspace{20mm}
\subcaptionbox{MSDM}{
\includegraphics[width=0.15\textwidth]{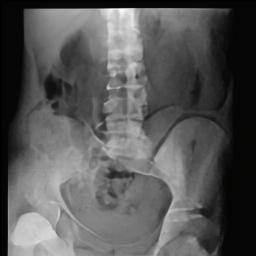} \includegraphics[width=0.15\textwidth]{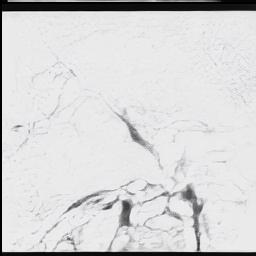} \includegraphics[width=0.15\textwidth]{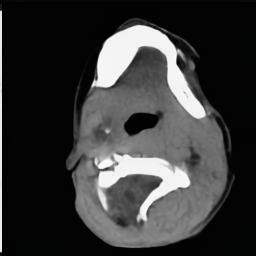} \includegraphics[width=0.15\textwidth]{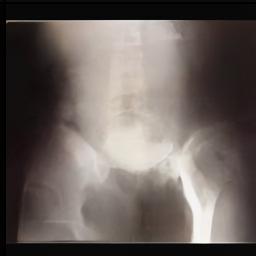} \includegraphics[width=0.15\textwidth]{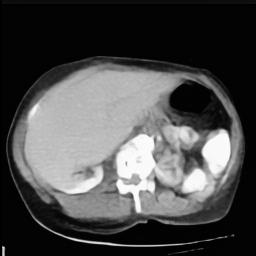}
\includegraphics[width=0.15\textwidth]{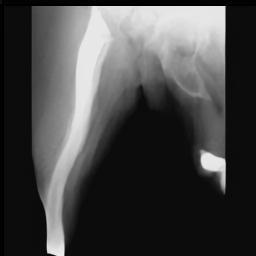}
}%
\caption{Examples of images generated using FLUX, Kandinsky 2.2 and MSDM compared to the original images from the ROCOv2 dataset.}
\label{fig:roco_models_pictures_comparison}
\end{figure}

We performed similar experiments on the ROCOv2 dataset and obtained the following FID scores for each model:

\begin{itemize}
    \item FLUX: 50.489 $\pm$ 1.09
    \item Kandinsky: 73.21 $\pm$ 4.08
    \item MSDM: 53.793 $\pm$ 2.65
\end{itemize}

The FID values were significantly lower on this dataset compared to the MedVQA-GI dataset. While it may not be completely valid to directly compare FID scores across different image modalities, we observed that the relative ranking of models remained consistent. However, the score variance was significantly lower than for models trained on MedVQA-GI. This indicates that the ROCOv2 dataset's larger size and broader anatomical coverage contributes to more stable learning signals.

Visual analysis of generated samples presented in Figure \ref{fig:roco_models_pictures_comparison} further confirms the models' strong performance on this dataset. The majority of generated images exhibit high realism and satisfactory detail preservation, demonstrating good generalization capabilities to non-GI medical imaging.

\subsubsection{Analysis of the impact of LoRa parameters}
\label{sec:results_3}

\begin{table*}[h!]
    \centering
    \caption{Kandinsky 2.2 and FLUX number of trainable parameters and FID scores (mean $\pm$ std over 5 runs) on private test dataset for different LoRA ranks.}
    \label{tab:lora_rank_results}
    \begin{tabular}{ lllll }
    \hline
        \thead{\textbf{LoRa} \\ \textbf{Rank}} & \thead{\textbf{Kandinsky 2.2} \\ \textbf{Param. count}} & \thead{\textbf{FLUX} \\ \textbf{Param. count}} & \thead{\textbf{Kandinsky 2.2} \\ \textbf{Mean FID} ($\downarrow$)} & \thead{\textbf{FLUX} \\ \textbf{Mean FID} ($\downarrow$)} \\ \hline
        4 & 2.1M & 11.2M & 116.95 $\pm$ 9.95 & 72.65 $\pm$ 3.43\\ 
        8 & 4.3M & 22.4M & 124.06 $\pm$ 10.72 & 70.11 $\pm$ 5.72\\ 
        16 & 8.5M & 44.8M & 120.24 $\pm$ 8.29 & 73.26 $\pm$ 2.79\\
        32 & 17.0M & 89.7M & 115.84 $\pm$ 8.24 & 71.33 $\pm$ 7.81\\ 
        64 & 34.1M & 179.3M & 112.38 $\pm$ 6.99 & 69.32 $\pm$ 6.18 \\
        128 & 68.3M & 358.6M & 124.82 $\pm$ 8.98 & 78,15 $\pm$ 9.14\\
        256 & 136.6M & 717.2M & 133.53 $\pm$ 8.73 & 75.91 $\pm$ 6.79\\ \hline
    \end{tabular}
\end{table*}

Given the central role of LoRA fine-tuning in our work, we conducted experiments to evaluate how its key hyperparameters affect generation quality. We focused primarily on two core parameters: rank and alpha.

Initial experiments with the alpha parameter, conducted with fixed rank values, demonstrated that adopting alpha equal to rank while optimizing the learning rate provided superior results. This aligns with LoRA's mathematical formulation, where alpha acts as a learning rate scaling factor.

To evaluate the impact of rank, we conducted extensive testing across rank values ranging from 4 to 256 for both Kandinsky 2.2 and FLUX models. Each configuration was evaluated five times on our private test set in the same way as in Section \ref{sec:results_1}. The results are presented in Table \ref{tab:lora_rank_results}, including mean FID scores, their standard deviations, and the corresponding number of trainable parameters for each rank configuration.

Our analysis revealed two key insights. First, FID scores showed significant instability across random seeds, although FLUX had significantly lower variance than Kandinsky 2.2, likely due to its larger scale and more sophisticated architecture. Second, our experiments demonstrated that rank selection had minimal impact on overall generation quality, suggesting that even modest rank values suffice for effective knowledge retention in contemporary large-scale models. This finding emphasizes the greater relative importance of training data quality. However, the higher ranks showed slightly more stable learning with slightly better quality (particularly rank 64, as shown in Table \ref{tab:lora_rank_results}), so we chose the final rank of 64 as a trade-off between size and stability.

Additional experiments with the ROCOv2 dataset further supported these findings. In the X-ray imaging modality with a significantly larger training data set, the LoRa ranking selection also showed a negligible impact on the FID metrics. The values observed remained within the ranges reported in Section \ref{sec:results_1}.

\begin{figure}[t!]
    \centering
    \includegraphics[width=0.95\linewidth]{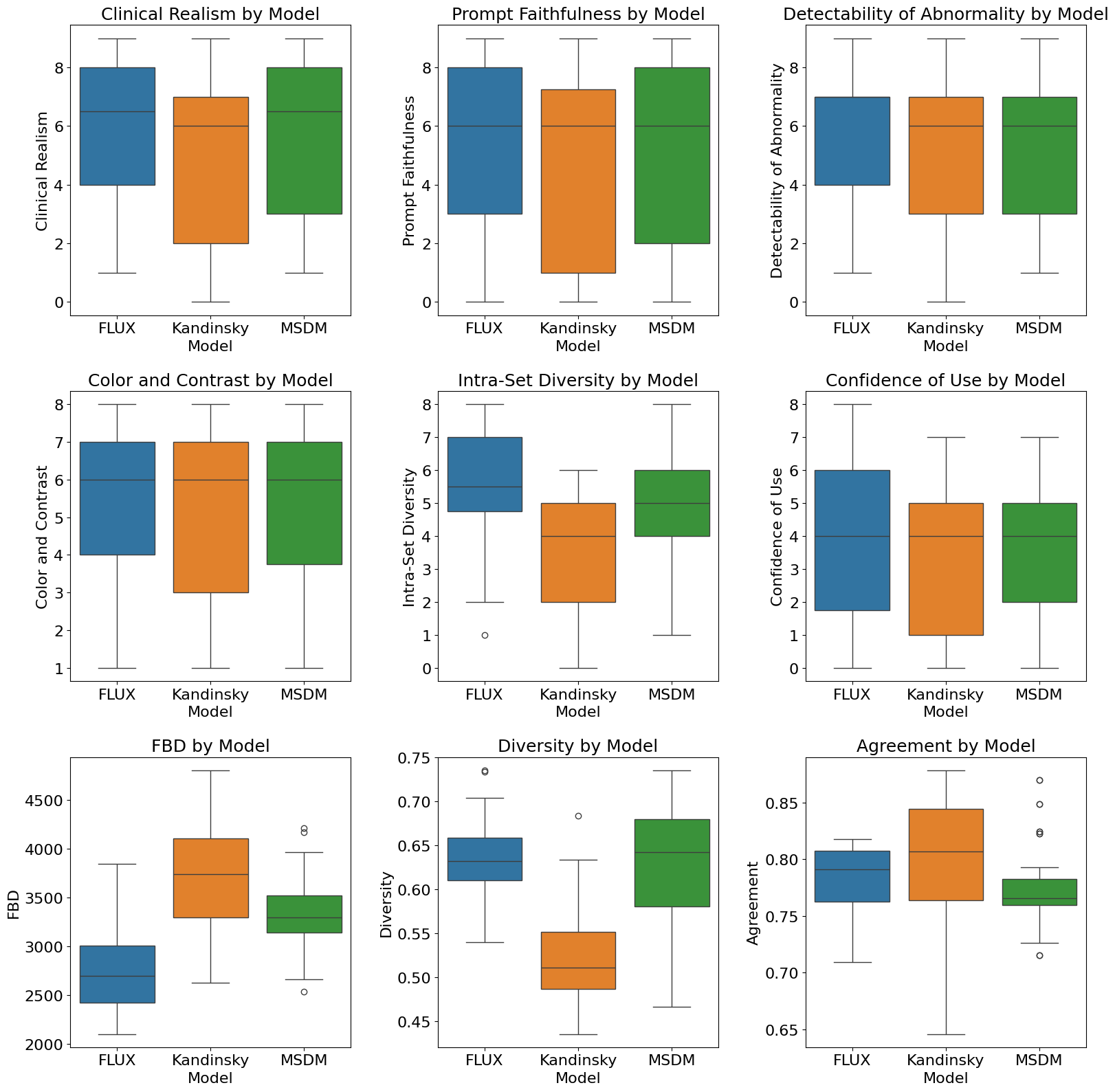}
    \caption{Human expert and quantitive metrics evaluation scores for each model across 40 test prompts.}
    \label{fig:human_metrics}
\end{figure}

\subsubsection{Expert and Quantitive Evaluation Results}

The full results of the expert medical assessment, along with FBD, Diversity, and Agreement scores, are presented in Figure \ref{fig:human_metrics}. We also measured Fidelity, with FLUX scoring 0.36, Kandinsky 2.2 scoring 0.26, and MSDM scoring 0.32 as shown in Table \ref{tab:model-comparison}. All models performed reasonably well in terms of prompt adherence, color/contrast, and general realism. However, their results differed more significantly across other metrics. In particular, the models' rankings based on human evaluation correlated strongly with their FID scores.
FLUX achieved the best—or comparable—performance across nearly all human-evaluated metrics, particularly in Intra-set Diversity and Confidence of Use. It also exhibited significantly lower FBD scores than competing models, further confirming the superior quality and realism of its generations. MSDM also performed competitively, often matching FLUX and consistently outperforming Kandinsky, even achieving the highest Diversity score among all models. In contrast, Kandinsky lagged behind, demonstrating notably weaker realism and the lowest generation diversity. Its sole advantage was in Agreement, though this likely stems from its limited variability—the model’s outputs align well across prompts and paraphrases precisely because it produces less diverse generations.

Despite the overall acceptable realism and quality, all metrics exhibited substantial variability, indicating that model performance is highly unstable. Generations can range from excellent to poor depending on random factors, which aligns with the previously observed high variance in FID scores. Additionally, the medical expert’s ratings reflected moderate confidence in the potential real-world applicability of these images. While this may partially reflect inherent skepticism toward AI-generated medical content, it nonetheless highlights the need for more consistent and clinically reliable solutions.

\begin{figure}[h!]
    \centering
    \includegraphics[width=1\linewidth]{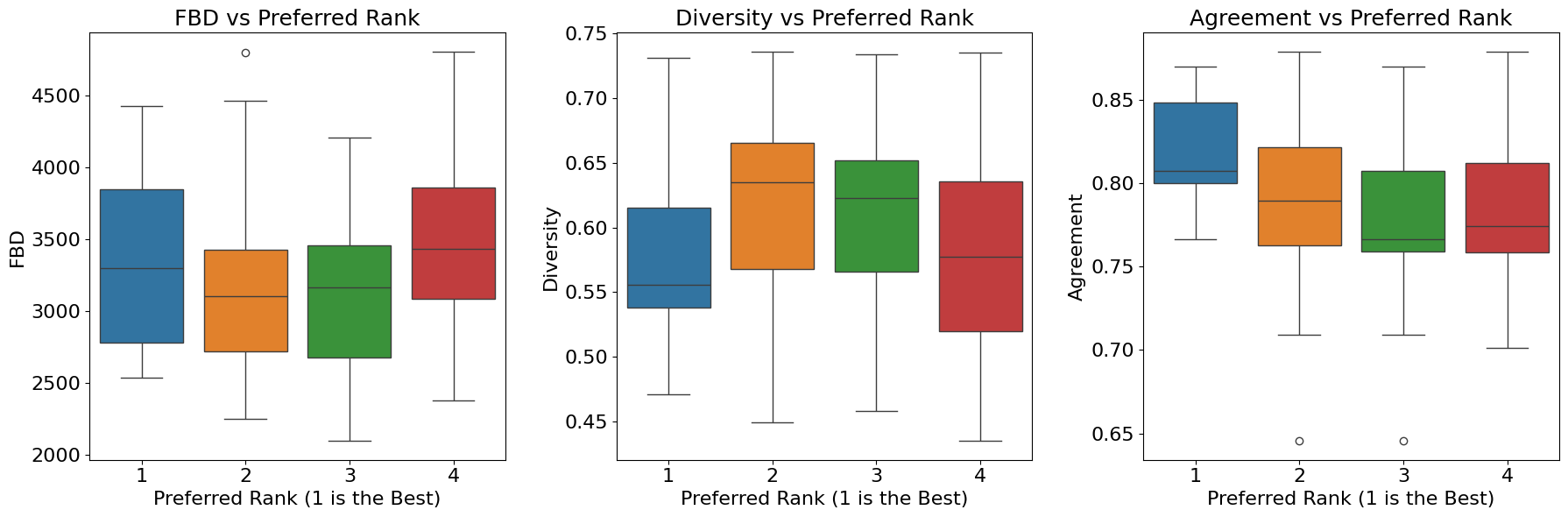}
    \caption{Correlation between automatic metrics (FBD, Diversity, Agreement) and human-expert rankings for each model (including real images) across all prompts.}
    \label{fig:metrics_vs_ranks}
\end{figure}

To evaluate the alignment between automatic metrics (FBD, Diversity, Agreement) and human assessments, we analyzed expert rankings across all prompts. The expert assigned each of the three models and real images a rank from 1 (best) to 4 for every prompt. These rankings were then compared against the average metric values per rank tier, as shown in Figure \ref{fig:metrics_vs_ranks}.
The analysis revealed several consistent patterns. Higher expert rankings strongly correlated with lower FBD scores (indicating better realism) and higher Agreement values (reflecting better prompt alignment). In contrast, Diversity showed a weaker relationship with rankings, suggesting it played a secondary role in human evaluation. These trends demonstrate that automatic metrics—particularly FBD for realism and Agreement for prompt fidelity—effectively capture the key factors driving human judgment.

\section{Discussion}

\label{sec:dics}

This study demonstrates the efficacy of diffusion models for controlled text-to-image generation in gastrointestinal endoscopy, with direct relevance to data augmentation, education, and privacy-preserving image synthesis. By combining parameter-efficient fine-tuning (LoRA) of large-scale models (e.g., Kandinsky 2.2, FLUX) with a lightweight, domain-adapted model (MSDM), we achieved state-of-the-art performance in the ImageCLEFmed 2024 MEDVQA-GI task. The resulting images reached near-photorealistic fidelity, with low FID scores, reflecting strong alignment between text prompts and generated anatomical content.
Our approach offers both flexibility and scalability in low-resource clinical settings. Fine-tuning billion-parameter models with LoRA enabled rapid adaptation to the limited MedVQA-GI dataset without prohibitive compute demands. Furthermore, MSDM—built on Stable Diffusion with medical-specific conditioning—highlighted the benefits of task-specific architecture in generating anatomically coherent outputs from textual descriptions.

\subsection{Human and Automated Evaluation}

To evaluate image quality and clinical relevance, we employed both automated metrics (e.g., FID) and human expert assessment. While FID offered a quantitative measure of distributional alignment, it failed to fully capture clinical nuance. Experts identified subtle diagnostic features—such as mucosal texture, vascularity, or polyp morphology—that were often overlooked by FID or similarity metrics alone.
Interestingly, some images with low FID lacked clinical utility due to anatomical inaccuracies, whereas others flagged as outliers by FID were deemed plausible by clinicians. This mismatch underscores the limitations of generic computer vision metrics in medical synthesis and emphasizes the need for domain-specific benchmarks grounded in clinical semantics.

\subsection{Limitations and Future Directions}

Although our approach demonstrates promising results in medical image synthesis, several important challenges remain. The models exhibit notable sensitivity to prompt specificity, while anatomically plausible outputs are consistently generated, diagnostically critical features can be imprecise when text inputs are underspecified or ambiguous. This limitation is compounded by the current restriction to static 2D image generation, which diverges from real-world clinical contexts like colonoscopy that rely on dynamic, multi-perspective video analysis.  

The MedVQA-GI dataset’s focus on common pathologies may introduce representation biases, potentially limiting model performance on rare but clinically significant findings. Furthermore, while quantitative metrics such as FID scores indicate strong generative performance, they do not fully capture diagnostic utility, a gap that requires continued expert validation. The inherent variability of natural language prompts presents additional challenges, suggesting that structured inputs or medical ontology-based conditioning could improve precision. Ethical considerations also emerge as synthetic images approach photorealism, requiring robust safeguards against misuse and clear protocols for traceability in clinical applications.  

Looking ahead, there are critical opportunities to advance the field through multimodal conditioning (incorporating radiomics, segmentation masks, or patient gaze patterns), bias-aware training frameworks, and interactive clinician-in-the-loop refinement systems. Moving beyond conventional metrics to task-specific validation, such as downstream classification or segmentation performance, will better assess clinical relevance. Future work should also explore practical integration pathways, from medical simulation training to privacy-preserving data sharing. These directions collectively aim to bridge the gap between technical capability and clinical adoption, ensuring that that diffusion models can meet the rigorous demands of real-world healthcare applications while maintaining ethical standards.

\section{Conclusion} %

\label{sec:conc}

Our evaluation demonstrates that optimized diffusion models can generate medically plausible images, with FLUX achieving the best performance (FID 72.96 on MedVQA-GI, 50.49 on ROCOv2) and our MSDM offering competitive quality at lower computational cost. While automated metrics correlated well with expert assessments, all models showed significant variance in output quality, particularly for subtle pathological features.

Key findings reveal that model architecture and dataset quality outweigh LoRA parameter choices in importance, and that larger, more diverse datasets (ROCOv2) enable better generalization. Current systems show promise for educational applications but require further refinement for diagnostic use.

For clinical adoption, we recommend: (1) hybrid evaluation combining metrics with expert review, (2) specialized benchmarks for diagnostic relevance, and (3) improved model stability. Future work should focus on pathological detail accuracy and clinician-in-the-loop refinement systems to bridge the gap between technical capability and clinical requirements.

\clearpage
\section*{Acknowledgements}
The work of E.T. was supported by the Russian Science Foundation grant \#23-11-00358. The study also utilized computational resources provided by the HPC facilities at HSE University \cite{cluster-link}. The contributions of S.G. and S.H. benefited from the Experimental Infrastructure for Exploration of Exascale Computing (eX3), financially supported by the Research Council of Norway under contract 270053.

We thank the medical personnel affiliated with the Nepali Army for their voluntary and enthusiastic assistance in annotating and validating expert evaluation data. Their clinical insights were instrumental in conducting the human assessment study. We also thank our anonymous medical expert for providing the independent evaluations used in the ranking and alignment analyses.

\section*{Author Contributions}

M.C. and E.T. led the study's conceptualization and contributed to implementing diffusion models. M.C. also handled LoRA fine-tuning, training on the HSE cluster, and assisted with data preparation and visualization. S.G. led the design of the evaluation framework, including human and automated metrics, developed the expert annotation interface, coordinated human evaluation, and conducted statistical analysis. He also contributed to visualization. Evaluation metrics were discussed and refined by all authors. S.H. supported dataset preparation for MedVQA-GI and enabled access to the test and private test set. All authors contributed to writing and editing the manuscript.

\section*{Use of AI Disclosure}

Various AI/LLM tools were used to draft structure, improve language and clarity. All content has been carefully manually reviewed, verified, and finalized by the authors.

\section*{Note to Readers}

In earlier draft of this manuscript, we used the term \textit{clinically-aware} to describe our approach to generating medical images from text prompts. We acknowledge that this terminology may have implied a level of clinical understanding that these models do not possess. The models presented in this work are not clinically aware—they are conditioned on structured medical text and trained to generate images that visually align with such prompts. 
In this version, we adopt the more accurate term \textit{medical text-conditioned} to reflect the model's capabilities without overstating their clinical interpretability or judgment.

\bibliography{main}

\begin{thebibliography}{10}
\expandafter\ifx\csname url\endcsname\relax
  \def\url#1{\texttt{#1}}\fi
\expandafter\ifx\csname urlprefix\endcsname\relax\def\urlprefix{URL }\fi
\expandafter\ifx\csname href\endcsname\relax
  \def\href#1#2{#2} \def\path#1{#1}\fi

\bibitem{kazerouni2023diffusion}
A.~Kazerouni, E.~K. Aghdam, M.~Heidari, R.~Azad, M.~Fayyaz, I.~Hacihaliloglu, D.~Merhof, {Diffusion models in medical imaging: A comprehensive survey}, Medical image analysis 88 (2023) 102846.

\bibitem{med-tasks-2}
A.~S. Lundervold, A.~Lundervold, \href{https://www.sciencedirect.com/science/article/pii/S0939388918301181}{{An overview of deep learning in medical imaging focusing on MRI}}, Zeitschrift für Medizinische Physik 29~(2) (2019) 102--127, special Issue: Deep Learning in Medical Physics.
\newblock \href {http://dx.doi.org/https://doi.org/10.1016/j.zemedi.2018.11.002} {\path{doi:https://doi.org/10.1016/j.zemedi.2018.11.002}}.
\newline\urlprefix\url{https://www.sciencedirect.com/science/article/pii/S0939388918301181}

\bibitem{med-tasks-3}
D.~Shen, G.~Wu, H.-I. Suk, \href{https://www.annualreviews.org/content/journals/10.1146/annurev-bioeng-071516-044442}{{Deep Learning in Medical Image Analysis}}, Annual Review of Biomedical Engineering 19~(Volume 19, 2017) (2017) 221--248.
\newblock \href {http://dx.doi.org/https://doi.org/10.1146/annurev-bioeng-071516-044442} {\path{doi:https://doi.org/10.1146/annurev-bioeng-071516-044442}}.
\newline\urlprefix\url{https://www.annualreviews.org/content/journals/10.1146/annurev-bioeng-071516-044442}

\bibitem{chan2020deep}
H.-P. Chan, R.~K. Samala, L.~M. Hadjiiski, C.~Zhou, {Deep learning in medical image analysis}, Deep learning in medical image analysis: challenges and applications (2020) 3--21.

\bibitem{pinaya2022brain}
W.~H. Pinaya, P.-D. Tudosiu, J.~Dafflon, P.~F. Da~Costa, V.~Fernandez, P.~Nachev, S.~Ourselin, M.~J. Cardoso, {Brain imaging generation with latent diffusion models}, in: MICCAI Workshop on Deep Generative Models, Springer, 2022, pp. 117--126.

\bibitem{willemink2020preparing}
M.~J. Willemink, W.~A. Koszek, C.~Hardell, J.~Wu, D.~Fleischmann, H.~Harvey, L.~R. Folio, R.~M. Summers, D.~L. Rubin, M.~P. Lungren, {Preparing medical imaging data for machine learning}, Radiology 295~(1) (2020) 4--15.

\bibitem{armanious2020medgan}
K.~Armanious, C.~Jiang, M.~Fischer, T.~K{\"u}stner, T.~Hepp, K.~Nikolaou, S.~Gatidis, B.~Yang, {MedGAN: Medical image translation using GANs}, Computerized medical imaging and graphics 79 (2020) 101684.

\bibitem{chung2022score}
H.~Chung, J.~C. Ye, {Score-based diffusion models for accelerated MRI}, Medical image analysis 80 (2022) 102479.

\bibitem{karimi2024diffusion}
D.~Karimi, S.~K. Warfield, {Diffusion mri with machine learning}, Imaging Neuroscience 2 (2024) 1--55.

\bibitem{kandinsky2_main_paper}
A.~Razzhigaev, A.~Shakhmatov, A.~Maltseva, V.~Arkhipkin, I.~Pavlov, I.~Ryabov, A.~Kuts, A.~Panchenko, A.~Kuznetsov, D.~Dimitrov, {Kandinsky: An Improved Text-to-Image Synthesis with Image Prior and Latent Diffusion}, in: Proceedings of the 2023 Conference on Empirical Methods in Natural Language Processing: System Demonstrations, 2023, pp. 286--295.

\bibitem{flux2024}
B.~F. Labs, {FLUX}, \url{https://github.com/black-forest-labs/flux} (2024).

\bibitem{diffusionpaper}
J.~Ho, A.~Jain, P.~Abbeel, \href{https://proceedings.neurips.cc/paper/2020/file/4c5bcfec8584af0d967f1ab10179ca4b-Paper.pdf}{{Denoising Diffusion Probabilistic Models}}, in: H.~Larochelle, M.~Ranzato, R.~Hadsell, M.~Balcan, H.~Lin (Eds.), Advances in Neural Information Processing Systems, Vol.~33, Curran Associates, Inc., 2020, pp. 6840--6851.
\newline\urlprefix\url{https://proceedings.neurips.cc/paper/2020/file/4c5bcfec8584af0d967f1ab10179ca4b-Paper.pdf}

\bibitem{Rombach}
R.~Rombach, A.~Blattmann, D.~Lorenz, P.~Esser, B.~Ommer, {High-Resolution Image Synthesis with Latent Diffusion Models}, in: {IEEE/CVF Conference on Computer Vision and Pattern Recognition (CVPR)}, IEEE, 2022, pp. 18--24.
\newblock \href {http://dx.doi.org/10.1109/CVPR52688.2022.01042} {\path{doi:10.1109/CVPR52688.2022.01042}}.

\bibitem{unet-original}
O.~Ronneberger, P.~Fischer, T.~Brox, {U-Net: Convolutional Networks for Biomedical Image Segmentation}, in: N.~Navab, J.~Hornegger, W.~M. Wells, A.~F. Frangi (Eds.), Medical Image Computing and Computer-Assisted Intervention -- MICCAI 2015, Springer International Publishing, Cham, 2015, pp. 234--241.

\bibitem{lora}
E.~J. Hu, Y.~Shen, P.~Wallis, Z.~Allen-Zhu, Y.~Li, S.~Wang, L.~Wang, W.~Chen, {Lora: Low-rank adaptation of large language models}, arXiv preprint arXiv:2106.09685.

\bibitem{hicks2023overview}
S.~Hicks, A.~M. Stor{\aa}s, P.~Halvorsen, T.~de~Lange, M.~Riegler, V.~Thambawita, {Overview of ImageCLEFmedical 2023-Medical Visual Question Answering for Gastrointestinal Tract.}, in: CLEF (Working Notes), 2023, pp. 1316--1327.

\bibitem{rocov2}
J.~R{\"u}ckert, L.~Bloch, R.~Br{\"u}ngel, A.~Idrissi-Yaghir, H.~Sch{\"a}fer, C.~S. Schmidt, S.~Koitka, O.~Pelka, A.~B. Abacha, A.~G.~Seco~de Herrera, et~al., {Rocov2: Radiology objects in context version 2, an updated multimodal image dataset}, Scientific Data 11~(1) (2024) 688.

\bibitem{chaichuk2024mmcp}
M.~Chaichuk, E.~Tutubalina, \href{https://www.semanticscholar.org/paper/271797644}{Mmcp team at imageclefmed medvqa-gi 2024 task: Diffusion models for text-to-image generation of colonoscopy images} (2024).
\newline\urlprefix\url{https://www.semanticscholar.org/paper/271797644}

\bibitem{Islam2024Feb}
S.~Islam, {\relax Md}.~T. Aziz, H.~R. Nabil, J.~R. Jim, M.~F. Mridha, {\relax Md}.~M. Kabir, {Generative Adversarial Networks (GANs) in Medical Imaging: Advancements, Applications, and Challenges}, IEEE Access 12 (2024) 35728--35753.
\newblock \href {http://dx.doi.org/10.1109/ACCESS.2024.3370848} {\path{doi:10.1109/ACCESS.2024.3370848}}.

\bibitem{Kazerouni2023Aug}
A.~Kazerouni, E.~K. Aghdam, M.~Heidari, R.~Azad, M.~Fayyaz, I.~Hacihaliloglu, D.~Merhof, {Diffusion models in medical imaging: A comprehensive survey}, Med. Image Anal. 88 (2023) 102846.
\newblock \href {http://dx.doi.org/10.1016/j.media.2023.102846} {\path{doi:10.1016/j.media.2023.102846}}.

\bibitem{Bengesi2024May}
S.~Bengesi, H.~El-Sayed, M.~D.~K. Sarker, Y.~Houkpati, J.~Irungu, T.~Oladunni, {Advancements in Generative AI: A Comprehensive Review of GANs, GPT, Autoencoders, Diffusion Model, and Transformers}, IEEE Access 12 (2024) 69812--69837.
\newblock \href {http://dx.doi.org/10.1109/ACCESS.2024.3397775} {\path{doi:10.1109/ACCESS.2024.3397775}}.

\bibitem{Cao2024Feb}
H.~Cao, C.~Tan, Z.~Gao, Y.~Xu, G.~Chen, P.-A. Heng, {A Survey on Generative Diffusion Models}, IEEE Trans. Knowl. Data Eng. 36~(7) (2024) 2814--2830.
\newblock \href {http://dx.doi.org/10.1109/TKDE.2024.3361474} {\path{doi:10.1109/TKDE.2024.3361474}}.

\bibitem{pmlr-v139-nichol21a}
A.~Q. Nichol, P.~Dhariwal, \href{https://proceedings.mlr.press/v139/nichol21a.html}{{Improved Denoising Diffusion Probabilistic Models}}, in: M.~Meila, T.~Zhang (Eds.), Proceedings of the 38th International Conference on Machine Learning, Vol. 139 of Proceedings of Machine Learning Research, PMLR, 2021, pp. 8162--8171.
\newline\urlprefix\url{https://proceedings.mlr.press/v139/nichol21a.html}

\bibitem{Hang2024Jul}
T.~Hang, S.~Gu, X.~Geng, B.~Guo, {Improved Noise Schedule for Diffusion Training}, arXiv\href {http://arxiv.org/abs/2407.03297} {\path{arXiv:2407.03297}}, \href {http://dx.doi.org/10.48550/arXiv.2407.03297} {\path{doi:10.48550/arXiv.2407.03297}}.

\bibitem{Choi_2022_CVPR}
J.~Choi, J.~Lee, C.~Shin, S.~Kim, H.~Kim, S.~Yoon, {Perception Prioritized Training of Diffusion Models}, in: Proceedings of the IEEE/CVF Conference on Computer Vision and Pattern Recognition (CVPR), 2022, pp. 11472--11481.

\bibitem{Chen2023Jan}
T.~Chen, {On the Importance of Noise Scheduling for Diffusion Models}, arXiv\href {http://arxiv.org/abs/2301.10972} {\path{arXiv:2301.10972}}, \href {http://dx.doi.org/10.48550/arXiv.2301.10972} {\path{doi:10.48550/arXiv.2301.10972}}.

\bibitem{Ho2022Jul}
J.~Ho, T.~Salimans, {Classifier-Free Diffusion Guidance}, arXiv\href {http://arxiv.org/abs/2207.12598} {\path{arXiv:2207.12598}}, \href {http://dx.doi.org/10.48550/arXiv.2207.12598} {\path{doi:10.48550/arXiv.2207.12598}}.

\bibitem{Preechakul2022}
K.~Preechakul, N.~Chatthee, S.~Wizadwongsa, S.~Suwajanakorn, \href{https://openaccess.thecvf.com/content/CVPR2022/html/Preechakul_Diffusion_Autoencoders_Toward_a_Meaningful_and_Decodable_Representation_CVPR_2022_paper.html}{{Diffusion Autoencoders: Toward a Meaningful and Decodable Representation}}, [Online; accessed 28. Apr. 2025] (2022).
\newline\urlprefix\url{https://openaccess.thecvf.com/content/CVPR2022/html/Preechakul_Diffusion_Autoencoders_Toward_a_Meaningful_and_Decodable_Representation_CVPR_2022_paper.html}

\bibitem{AllWorthWords}
F.~Bao, S.~Nie, K.~Xue, Y.~Cao, C.~Li, H.~Su, {All are Worth Words: A ViT Backbone for Diffusion Models}, in: {IEEE/CVF Conference on Computer Vision and Pattern Recognition (CVPR)}, IEEE, 2023, pp. 17--24.
\newblock \href {http://dx.doi.org/10.1109/CVPR52729.2023.02171} {\path{doi:10.1109/CVPR52729.2023.02171}}.

\bibitem{Saxena2021May}
D.~Saxena, J.~Cao, {Generative Adversarial Networks (GANs): Challenges, Solutions, and Future Directions}, ACM Comput. Surv. 54~(3) (2021) 1--42.
\newblock \href {http://dx.doi.org/10.1145/3446374} {\path{doi:10.1145/3446374}}.

\bibitem{Muller-Franzes2023Jul}
G.~M{\ifmmode\ddot{u}\else\"{u}\fi}ller-Franzes, J.~M. Niehues, F.~Khader, S.~T. Arasteh, C.~Haarburger, C.~Kuhl, T.~Wang, T.~Han, T.~Nolte, S.~Nebelung, et~al., {A multimodal comparison of latent denoising diffusion probabilistic models and generative adversarial networks for medical image synthesis}, Sci. Rep. 13 (2023) 12098.
\newblock \href {http://dx.doi.org/10.1038/s41598-023-39278-0} {\path{doi:10.1038/s41598-023-39278-0}}.

\bibitem{DiffuseVAE}
K.~Pandey, A.~Mukherjee, P.~Rai, A.~Kumar, \href{https://arxiv.org/abs/2201.00308}{{DiffuseVAE: Efficient, Controllable and High-Fidelity Generation from Low Dimensional Latents}}, TMLR.
\newline\urlprefix\url{https://arxiv.org/abs/2201.00308}

\bibitem{Sadat2024May}
S.~Sadat, J.~Buhmann, D.~Bradley, O.~Hilliges, R.~M. Weber, {LiteVAE: Lightweight and Efficient Variational Autoencoders for Latent Diffusion Models}, arXiv\href {http://arxiv.org/abs/2405.14477} {\path{arXiv:2405.14477}}, \href {http://dx.doi.org/10.48550/arXiv.2405.14477} {\path{doi:10.48550/arXiv.2405.14477}}.

\bibitem{He2025Feb}
C.~He, Y.~Shen, C.~Fang, F.~Xiao, L.~Tang, Y.~Zhang, {Diffusion Models in Low-Level Vision: A Survey}, IEEE Trans. Pattern Anal. Mach. Intell. (2025) 1--20\href {http://dx.doi.org/10.1109/TPAMI.2025.3545047} {\path{doi:10.1109/TPAMI.2025.3545047}}.

\bibitem{Zhang_2017_ICCV}
H.~Zhang, T.~Xu, H.~Li, S.~Zhang, X.~Wang, X.~Huang, D.~Metaxas, {StackGAN: Text to Photo-Realistic Image Synthesis with Stacked Generative Adversarial Networks}, IEEE Computer Society, 2017.
\newblock \href {http://dx.doi.org/10.1109/ICCV.2017.629} {\path{doi:10.1109/ICCV.2017.629}}.

\bibitem{Xu_2018_CVPR}
T.~Xu, P.~Zhang, Q.~Huang, H.~Zhang, Z.~Gan, X.~Huang, X.~He, {AttnGAN: Fine-Grained Text to Image Generation with Attentional Generative Adversarial Networks}, IEEE Computer Society, 2018.
\newblock \href {http://dx.doi.org/10.1109/CVPR.2018.00143} {\path{doi:10.1109/CVPR.2018.00143}}.

\bibitem{NEURIPS2019_5f8e2fa1}
A.~Razavi, A.~van~den Oord, O.~Vinyals, {Generating diverse high-fidelity images with VQ-VAE-2}, in: {Guide Proceedings}, Curran Associates Inc., 2019, pp. 14866--14876.
\newblock \href {http://dx.doi.org/10.5555/3454287.3455618} {\path{doi:10.5555/3454287.3455618}}.

\bibitem{pmlr-v139-ramesh21a}
A.~Ramesh, M.~Pavlov, G.~Goh, S.~Gray, C.~Voss, A.~Radford, M.~Chen, I.~Sutskever, \href{https://proceedings.mlr.press/v139/ramesh21a.html}{{Zero-Shot Text-to-Image Generation}}, in: M.~Meila, T.~Zhang (Eds.), Proceedings of the 38th International Conference on Machine Learning, Vol. 139 of Proceedings of Machine Learning Research, PMLR, 2021, pp. 8821--8831.
\newline\urlprefix\url{https://proceedings.mlr.press/v139/ramesh21a.html}

\bibitem{Nichol2021Dec}
A.~Nichol, P.~Dhariwal, A.~Ramesh, P.~Shyam, P.~Mishkin, B.~McGrew, I.~Sutskever, M.~Chen, {GLIDE: Towards Photorealistic Image Generation and Editing with Text-Guided Diffusion Models}, arXiv\href {http://arxiv.org/abs/2112.10741} {\path{arXiv:2112.10741}}, \href {http://dx.doi.org/10.48550/arXiv.2112.10741} {\path{doi:10.48550/arXiv.2112.10741}}.

\bibitem{Ramesh2022Apr}
A.~Ramesh, P.~Dhariwal, A.~Nichol, C.~Chu, M.~Chen, {Hierarchical Text-Conditional Image Generation with CLIP Latents}, arXiv\href {http://arxiv.org/abs/2204.06125} {\path{arXiv:2204.06125}}, \href {http://dx.doi.org/10.48550/arXiv.2204.06125} {\path{doi:10.48550/arXiv.2204.06125}}.

\bibitem{Saharia2022May}
C.~Saharia, W.~Chan, S.~Saxena, L.~Li, J.~Whang, E.~Denton, S.~K.~S. Ghasemipour, B.~K. Ayan, S.~S. Mahdavi, R.~G. Lopes, et~al., {Photorealistic Text-to-Image Diffusion Models with Deep Language Understanding}, arXiv\href {http://arxiv.org/abs/2205.11487} {\path{arXiv:2205.11487}}, \href {http://dx.doi.org/10.48550/arXiv.2205.11487} {\path{doi:10.48550/arXiv.2205.11487}}.

\bibitem{Razzhigaev2023Oct}
A.~Razzhigaev, A.~Shakhmatov, A.~Maltseva, V.~Arkhipkin, I.~Pavlov, I.~Ryabov, A.~Kuts, A.~Panchenko, A.~Kuznetsov, D.~Dimitrov, {Kandinsky: an Improved Text-to-Image Synthesis with Image Prior and Latent Diffusion}, arXiv\href {http://arxiv.org/abs/2310.03502} {\path{arXiv:2310.03502}}, \href {http://dx.doi.org/10.48550/arXiv.2310.03502} {\path{doi:10.48550/arXiv.2310.03502}}.

\bibitem{Zhang2024May}
X.~Zhang, X.~Wei, W.~Hu, J.~Wu, J.~Wu, W.~Zhang, Z.~Zhang, Z.~Lei, Q.~Li, {A Survey on Personalized Content Synthesis with Diffusion Models}, arXiv\href {http://arxiv.org/abs/2405.05538} {\path{arXiv:2405.05538}}, \href {http://dx.doi.org/10.48550/arXiv.2405.05538} {\path{doi:10.48550/arXiv.2405.05538}}.

\bibitem{dalle2Radiologoy}
{What Does DALL-E 2 Know About Radiology?}, [Online; accessed 28. Apr. 2025] (Jan. 2023).
\newblock \href {http://dx.doi.org/10.2196/43110} {\path{doi:10.2196/43110}}.

\bibitem{Altaf2019Jul}
F.~Altaf, S.~M.~S. Islam, N.~Akhtar, N.~K. Janjua, {Going Deep in Medical Image Analysis: Concepts, Methods, Challenges, and Future Directions}, IEEE Access 7 (2019) 99540--99572.
\newblock \href {http://dx.doi.org/10.1109/ACCESS.2019.2929365} {\path{doi:10.1109/ACCESS.2019.2929365}}.

\bibitem{Nazir2024Jul}
A.~Nazir, A.~Hussain, M.~Singh, A.~Assad, {Deep learning in medicine: advancing healthcare with intelligent solutions and the future of holography imaging in early diagnosis}, Multimed. Tools Appl. (2024) 1--64\href {http://dx.doi.org/10.1007/s11042-024-19694-8} {\path{doi:10.1007/s11042-024-19694-8}}.

\bibitem{Panayides2020May}
A.~S. Panayides, A.~Amini, N.~D. Filipovic, A.~Sharma, S.~A. Tsaftaris, A.~Young, {AI in Medical Imaging Informatics: Current Challenges and Future Directions}, IEEE J. Biomed. Health Inf. 24~(7) (2020) 1837--1857.
\newblock \href {http://dx.doi.org/10.1109/JBHI.2020.2991043} {\path{doi:10.1109/JBHI.2020.2991043}}.

\bibitem{Najjar2023Aug}
R.~Najjar, {Redefining Radiology: A Review of Artificial Intelligence Integration in Medical Imaging}, Diagnostics 13~(17) (2023) 2760.
\newblock \href {http://dx.doi.org/10.3390/diagnostics13172760} {\path{doi:10.3390/diagnostics13172760}}.

\bibitem{Zhang2024Jan}
S.~Zhang, D.~Metaxas, {On the challenges and perspectives of foundation models for medical image analysis}, Med. Image Anal. 91 (2024) 102996.
\newblock \href {http://dx.doi.org/10.1016/j.media.2023.102996} {\path{doi:10.1016/j.media.2023.102996}}.

\bibitem{Khan2025Jan}
W.~Khan, S.~Leem, K.~B. See, J.~K. Wong, S.~Zhang, R.~Fang, {A Comprehensive Survey of Foundation Models in Medicine}, IEEE Rev. Biomed. Eng. (2025) 1--20\href {http://dx.doi.org/10.1109/RBME.2025.3531360} {\path{doi:10.1109/RBME.2025.3531360}}.

\bibitem{Alsentzer2019Apr}
E.~Alsentzer, J.~R. Murphy, W.~Boag, W.-H. Weng, D.~Jin, T.~Naumann, M.~B.~A. McDermott, {Publicly Available Clinical BERT Embeddings}, arXiv\href {http://arxiv.org/abs/1904.03323} {\path{arXiv:1904.03323}}, \href {http://dx.doi.org/10.48550/arXiv.1904.03323} {\path{doi:10.48550/arXiv.1904.03323}}.

\bibitem{Yan2022Jun}
A.~Yan, J.~McAuley, X.~Lu, J.~Du, E.~Y. Chang, A.~Gentili, C.-N. Hsu, \href{https://pubs.rsna.org/doi/full/10.1148/ryai.210258}{{RadBERT: Adapting Transformer-based Language Models to Radiology}}, Radiology: Artificial Intelligence.
\newline\urlprefix\url{https://pubs.rsna.org/doi/full/10.1148/ryai.210258}

\bibitem{Lozano2025Jan}
A.~Lozano, M.~W. Sun, J.~Burgess, L.~Chen, J.~J. Nirschl, J.~Gu, I.~Lopez, J.~Aklilu, A.~W. Katzer, C.~Chiu, et~al., {BIOMEDICA: An Open Biomedical Image-Caption Archive, Dataset, and Vision-Language Models Derived from Scientific Literature}, arXiv\href {http://arxiv.org/abs/2501.07171} {\path{arXiv:2501.07171}}, \href {http://dx.doi.org/10.48550/arXiv.2501.07171} {\path{doi:10.48550/arXiv.2501.07171}}.

\bibitem{Wang2025Feb}
J.~Wang, K.~Wang, Y.~Yu, Y.~Lu, W.~Xiao, Z.~Sun, F.~Liu, Z.~Zou, Y.~Gao, L.~Yang, et~al., {Self-improving generative foundation model for synthetic medical image generation and clinical applications}, Nat. Med. 31~(2) (2025) 609--617.
\newblock \href {http://dx.doi.org/10.1038/s41591-024-03359-y} {\path{doi:10.1038/s41591-024-03359-y}}.

\bibitem{Chambon2022Nov}
P.~Chambon, C.~Bluethgen, J.-B. Delbrouck, R.~Van~der Sluijs, M.~Po{\l}acin, J.~M.~Z. Chaves, T.~M. Abraham, S.~Purohit, C.~P. Langlotz, A.~Chaudhari, {RoentGen: Vision-Language Foundation Model for Chest X-ray Generation}, arXiv\href {http://arxiv.org/abs/2211.12737} {\path{arXiv:2211.12737}}, \href {http://dx.doi.org/10.48550/arXiv.2211.12737} {\path{doi:10.48550/arXiv.2211.12737}}.

\bibitem{Huang2024}
P.~Huang, X.~Gao, L.~Huang, J.~Jiao, X.~Li, Y.~Wang, {Chest-Diffusion: A Light-Weight Text-to-Image Model for Report-to-CXR Generation}, in: {IEEE International Symposium on Biomedical Imaging (ISBI)}, IEEE, 2024, pp. 27--30.
\newblock \href {http://dx.doi.org/10.1109/ISBI56570.2024.10635417} {\path{doi:10.1109/ISBI56570.2024.10635417}}.

\bibitem{Han2024Mar}
W.~Han, C.~Kim, D.~Ju, Y.~Shim, S.~J. Hwang, {Advancing Text-Driven Chest X-Ray Generation with Policy-Based Reinforcement Learning}, arXiv\href {http://arxiv.org/abs/2403.06516} {\path{arXiv:2403.06516}}, \href {http://dx.doi.org/10.48550/arXiv.2403.06516} {\path{doi:10.48550/arXiv.2403.06516}}.

\bibitem{Khader2023May}
F.~Khader, G.~M{\ifmmode\ddot{u}\else\"{u}\fi}ller-Franzes, S.~T. Arasteh, T.~Han, C.~Haarburger, M.~Schulze-Hagen, P.~Schad, S.~Engelhardt, B.~Bae{\ss}ler, S.~Foersch, et~al., {Denoising diffusion probabilistic models for 3D medical image generation}, Sci. Rep. 13~(1) (2023) 7303.
\newblock \href {http://arxiv.org/abs/37147413} {\path{arXiv:37147413}}, \href {http://dx.doi.org/10.1038/s41598-023-34341-2} {\path{doi:10.1038/s41598-023-34341-2}}.

\bibitem{Xu2023Oct}
Y.~Xu, L.~Sun, W.~Peng, S.~Jia, K.~Morrison, A.~Perer, A.~Zandifar, S.~Visweswaran, M.~Eslami, K.~Batmanghelich, {MedSyn: Text-guided Anatomy-aware Synthesis of High-Fidelity 3D CT Images}, arXiv\href {http://arxiv.org/abs/2310.03559} {\path{arXiv:2310.03559}}, \href {http://dx.doi.org/10.48550/arXiv.2310.03559} {\path{doi:10.48550/arXiv.2310.03559}}.

\bibitem{Pan2024Apr}
S.~Pan, E.~Abouei, J.~Wynne, C.-W. Chang, T.~Wang, R.~L.~J. Qiu, Y.~Li, J.~Peng, J.~Roper, P.~Patel, et~al., {Synthetic CT generation from MRI using 3D transformer-based denoising diffusion model}, Med. Phys. 51~(4) (2024) 2538--2548.
\newblock \href {http://arxiv.org/abs/38011588} {\path{arXiv:38011588}}, \href {http://dx.doi.org/10.1002/mp.16847} {\path{doi:10.1002/mp.16847}}.

\bibitem{Selim2023Jan}
M.~Selim, J.~Zhang, M.~A. Brooks, G.~Wang, J.~Chen, {DiffusionCT: Latent Diffusion Model for CT Image Standardization}, arXiv\href {http://arxiv.org/abs/2301.08815} {\path{arXiv:2301.08815}}, \href {http://dx.doi.org/10.48550/arXiv.2301.08815} {\path{doi:10.48550/arXiv.2301.08815}}.

\bibitem{Machacek2023Jun}
R.~Mach{\ifmmode\acute{a}\else\'{a}\fi}{\ifmmode\check{c}\else\v{c}\fi}ek, L.~Mozaffari, Z.~Sepasdar, S.~Parasa, P.~Halvorsen, M.~A. Riegler, V.~Thambawita, {Mask-conditioned latent diffusion for generating gastrointestinal polyp images}, in: {ACM Conferences}, Association for Computing Machinery, New York, NY, USA, 2023, pp. 1--9.
\newblock \href {http://dx.doi.org/10.1145/3592571.3592978} {\path{doi:10.1145/3592571.3592978}}.

\bibitem{Pishva}
A.~K. Pishva, V.~Thambawita, J.~Torresen, S.~A. Hicks, {RePolyp: A Framework for Generating Realistic Colon Polyps with Corresponding Segmentation Masks using Diffusion Models}, in: {IEEE 36th International Symposium on Computer-Based Medical Systems (CBMS)}, IEEE, 2023, pp. 22--24.
\newblock \href {http://dx.doi.org/10.1109/CBMS58004.2023.00190} {\path{doi:10.1109/CBMS58004.2023.00190}}.

\bibitem{Sharma2025Feb}
V.~Sharma, D.~Jha, M.~K. Bhuyan, P.~K. Das, U.~Bagci, {Diverse Image Generation with Diffusion Models and Cross Class Label Learning for Polyp Classification}, arXiv\href {http://arxiv.org/abs/2502.05444} {\path{arXiv:2502.05444}}, \href {http://dx.doi.org/10.48550/arXiv.2502.05444} {\path{doi:10.48550/arXiv.2502.05444}}.

\bibitem{Du2023Oct}
Y.~Du, Y.~Jiang, S.~Tan, X.~Wu, Q.~Dou, Z.~Li, G.~Li, X.~Wan, {ArSDM: Colonoscopy Images Synthesis with Adaptive Refinement Semantic Diffusion Models}, in: {Medical Image Computing and Computer Assisted Intervention {\textendash} MICCAI 2023}, Springer, Cham, Switzerland, 2023, pp. 339--349.
\newblock \href {http://dx.doi.org/10.1007/978-3-031-43895-0_32} {\path{doi:10.1007/978-3-031-43895-0_32}}.

\bibitem{Ionescu2024Sep}
B.~Ionescu, H.~M{\ifmmode\ddot{u}\else\"{u}\fi}ller, A.-M. Dr{\ifmmode\u{a}\else\u{a}\fi}gulinescu, J.~R{\ifmmode\ddot{u}\else\"{u}\fi}ckert, A.~Ben~Abacha, A.~Garc{\ifmmode\acute{\imath}\else\'{\i}\fi}a Seco~de Herrera, L.~Bloch, R.~Br{\ifmmode\ddot{u}\else\"{u}\fi}ngel, A.~Idrissi-Yaghir, H.~Sch{\ifmmode\ddot{a}\else\"{a}\fi}fer, et~al., {Overview of the ImageCLEF 2024: Multimedia Retrieval in Medical Applications}, in: {Experimental IR Meets Multilinguality, Multimodality, and Interaction}, Springer, Cham, Switzerland, 2024, pp. 140--164.
\newblock \href {http://dx.doi.org/10.1007/978-3-031-71908-0_7} {\path{doi:10.1007/978-3-031-71908-0_7}}.

\bibitem{Allmendinger2024Aug}
S.~Allmendinger, P.~Hemmer, M.~Queisner, I.~Sauer, L.~M{\ifmmode\ddot{u}\else\"{u}\fi}ller, J.~Jakubik, M.~V{\ifmmode\ddot{o}\else\"{o}\fi}ssing, N.~K{\ifmmode\ddot{u}\else\"{u}\fi}hl, {Navigating the Synthetic Realm: Harnessing Diffusion-Based Models for Laparoscopic Text-to-Image Generation}, in: {AI for Health Equity and Fairness: Leveraging AI to Address Social Determinants of Health}, Springer, Cham, Switzerland, 2024, pp. 31--46.
\newblock \href {http://dx.doi.org/10.1007/978-3-031-63592-2_4} {\path{doi:10.1007/978-3-031-63592-2_4}}.

\bibitem{peft}
S.~Mangrulkar, S.~Gugger, L.~Debut, Y.~Belkada, S.~Paul, B.~Bossan, {PEFT: State-of-the-art Parameter-Efficient Fine-Tuning methods}, \url{https://github.com/huggingface/peft} (2022).

\bibitem{Hu2021Jun}
E.~J. Hu, Y.~Shen, P.~Wallis, Z.~Allen-Zhu, Y.~Li, S.~Wang, L.~Wang, W.~Chen, {LoRA: Low-Rank Adaptation of Large Language Models}, arXiv\href {http://arxiv.org/abs/2106.09685} {\path{arXiv:2106.09685}}, \href {http://dx.doi.org/10.48550/arXiv.2106.09685} {\path{doi:10.48550/arXiv.2106.09685}}.

\bibitem{Mao2024Aug}
Y.~Mao, H.~Li, W.~Pang, G.~Papanastasiou, G.~Yang, C.~Wang, {SeLoRA: Self-Expanding Low-Rank Adaptation of Latent Diffusion Model for Medical Image Synthesis}, arXiv\href {http://arxiv.org/abs/2408.07196} {\path{arXiv:2408.07196}}, \href {http://dx.doi.org/10.48550/arXiv.2408.07196} {\path{doi:10.48550/arXiv.2408.07196}}.

\bibitem{Gal2022Aug}
R.~Gal, Y.~Alaluf, Y.~Atzmon, O.~Patashnik, A.~H. Bermano, G.~Chechik, D.~Cohen-Or, {An Image is Worth One Word: Personalizing Text-to-Image Generation using Textual Inversion}, arXiv\href {http://arxiv.org/abs/2208.01618} {\path{arXiv:2208.01618}}, \href {http://dx.doi.org/10.48550/arXiv.2208.01618} {\path{doi:10.48550/arXiv.2208.01618}}.

\bibitem{deWilde2023Mar}
B.~de~Wilde, A.~Saha, M.~de~Rooij, H.~Huisman, G.~Litjens, {Medical diffusion on a budget: Textual Inversion for medical image generation}, arXiv\href {http://arxiv.org/abs/2303.13430} {\path{arXiv:2303.13430}}, \href {http://dx.doi.org/10.48550/arXiv.2303.13430} {\path{doi:10.48550/arXiv.2303.13430}}.

\bibitem{Ruiz2022Aug}
N.~Ruiz, Y.~Li, V.~Jampani, Y.~Pritch, M.~Rubinstein, K.~Aberman, {DreamBooth: Fine Tuning Text-to-Image Diffusion Models for Subject-Driven Generation}, arXiv\href {http://arxiv.org/abs/2208.12242} {\path{arXiv:2208.12242}}, \href {http://dx.doi.org/10.48550/arXiv.2208.12242} {\path{doi:10.48550/arXiv.2208.12242}}.

\bibitem{Zhang2025Jan}
D.~Zhang, T.~Feng, L.~Xue, Y.~Wang, Y.~Dong, J.~Tang, {Parameter-Efficient Fine-Tuning for Foundation Models}, arXiv\href {http://arxiv.org/abs/2501.13787} {\path{arXiv:2501.13787}}, \href {http://dx.doi.org/10.48550/arXiv.2501.13787} {\path{doi:10.48550/arXiv.2501.13787}}.

\bibitem{Christophe2024Apr}
C.~Christophe, P.~K. Kanithi, P.~Munjal, T.~Raha, N.~Hayat, R.~Rajan, A.~Al-Mahrooqi, A.~Gupta, M.~U. Salman, G.~Gosal, et~al., {Med42 -- Evaluating Fine-Tuning Strategies for Medical LLMs: Full-Parameter vs. Parameter-Efficient Approaches}, arXiv\href {http://arxiv.org/abs/2404.14779} {\path{arXiv:2404.14779}}, \href {http://dx.doi.org/10.48550/arXiv.2404.14779} {\path{doi:10.48550/arXiv.2404.14779}}.

\bibitem{Heusel2017Jun}
M.~Heusel, H.~Ramsauer, T.~Unterthiner, B.~Nessler, S.~Hochreiter, {GANs Trained by a Two Time-Scale Update Rule Converge to a Local Nash Equilibrium}, arXiv\href {http://arxiv.org/abs/1706.08500} {\path{arXiv:1706.08500}}, \href {http://dx.doi.org/10.48550/arXiv.1706.08500} {\path{doi:10.48550/arXiv.1706.08500}}.

\bibitem{Woodland2023Nov}
M.~Woodland, A.~Castelo, M.~A. Taie, J.~A.~M. Silva, M.~Eltaher, F.~Mohn, A.~Shieh, S.~Kundu, J.~P. Yung, A.~B. Patel, et~al., {Feature Extraction for Generative Medical Imaging Evaluation: New Evidence Against an Evolving Trend}, arXiv\href {http://arxiv.org/abs/2311.13717} {\path{arXiv:2311.13717}}, \href {http://dx.doi.org/10.1007/978-3-031-72390-2_9} {\path{doi:10.1007/978-3-031-72390-2_9}}.

\bibitem{Cepa2024Sep}
B.~Cepa, C.~Brito, A.~Sousa, \href{https://doi.org/10.1101/2024.09.27.615343}{{To FID or not to FID: Applying GANs for MRI Image Generation in HPC}}, bioRxiv (2024) 2024.09.27.615343\href {http://arxiv.org/abs/2024.09.27.615343} {\path{arXiv:2024.09.27.615343}}.
\newline\urlprefix\url{https://doi.org/10.1101/2024.09.27.615343}

\bibitem{Sudre2017Jul}
C.~H. Sudre, W.~Li, T.~Vercauteren, S.~Ourselin, M.~J. Cardoso, {Generalised Dice overlap as a deep learning loss function for highly unbalanced segmentations}, arXiv\href {http://arxiv.org/abs/1707.03237} {\path{arXiv:1707.03237}}, \href {http://dx.doi.org/10.1007/978-3-319-67558-9_28} {\path{doi:10.1007/978-3-319-67558-9_28}}.

\bibitem{Zhang2023Oct}
Z.~Zhang, L.~Yao, B.~Wang, D.~Jha, G.~Durak, E.~Keles, A.~Medetalibeyoglu, U.~Bagci, {DiffBoost: Enhancing Medical Image Segmentation via Text-Guided Diffusion Model}, arXiv\href {http://arxiv.org/abs/2310.12868} {\path{arXiv:2310.12868}}, \href {http://dx.doi.org/10.48550/arXiv.2310.12868} {\path{doi:10.48550/arXiv.2310.12868}}.

\bibitem{Borgli2020}
H.~Borgli, V.~Thambawita, P.~H. Smedsrud, S.~Hicks, D.~Jha, S.~L. Eskeland, K.~R. Randel, K.~Pogorelov, M.~Lux, D.~T.~D. Nguyen, D.~Johansen, C.~Griwodz, H.~K. Stensland, E.~Garcia-Ceja, P.~T. Schmidt, H.~L. Hammer, M.~A. Riegler, P.~Halvorsen, T.~de~Lange, \href{https://doi.org/10.1038/s41597-020-00622-y}{{HyperKvasir, a comprehensive multi-class image and video dataset for gastrointestinal endoscopy}}, Scientific Data 7~(1) (2020) 283.
\newblock \href {http://dx.doi.org/10.1038/s41597-020-00622-y} {\path{doi:10.1038/s41597-020-00622-y}}.
\newline\urlprefix\url{https://doi.org/10.1038/s41597-020-00622-y}

\bibitem{jha2021kvasir}
D.~Jha, S.~Ali, K.~Emanuelsen, S.~A. Hicks, V.~Thambawita, E.~Garcia-Ceja, M.~A. Riegler, T.~d. Lange, P.~T. Schmidt, H.~D. Johansen, et~al., Kvasir-instrument: Diagnostic and therapeutic tool segmentation dataset in gastrointestinal endoscopy, in: Proceedings of the International Conference on Multimedia Modeling, Springer, 2021, pp. 218--229.

\bibitem{Gautam2024Oct}
S.~Gautam, A.~M. Stor{\aa}s, C.~Midoglu, S.~A. Hicks, V.~Thambawita, P.~Halvorsen, M.~A. Riegler, {Kvasir-VQA: A Text-Image Pair GI Tract Dataset}, in: {ACM Conferences}, Association for Computing Machinery, New York, NY, USA, 2024, pp. 3--12.
\newblock \href {http://dx.doi.org/10.1145/3689096.3689458} {\path{doi:10.1145/3689096.3689458}}.

\bibitem{kandinsky-top}
A.~Raza, \href{https://techbullion.com/a-showdown-of-creativity-a-comparative-analysis-of-proprietary-generative-ai-image-models/}{{TECHNOLOGYA Showdown of Creativity: A Comparative Analysis of Proprietary Generative AI Image Models}} (2024).
\newline\urlprefix\url{https://techbullion.com/a-showdown-of-creativity-a-comparative-analysis-of-proprietary-generative-ai-image-models/}

\bibitem{diffusion-mapping}
W.~Peebles, S.~Xie, {Scalable diffusion models with transformers}, in: Proceedings of the IEEE/CVF International Conference on Computer Vision, 2023, pp. 4195--4205.

\bibitem{clip}
A.~Radford, J.~W. Kim, C.~Hallacy, A.~Ramesh, G.~Goh, S.~Agarwal, G.~Sastry, A.~Askell, P.~Mishkin, J.~Clark, et~al., {Learning transferable visual models from natural language supervision}, in: International conference on machine learning, PMLR, 2021, pp. 8748--8763.

\bibitem{vqgan}
P.~Esser, R.~Rombach, B.~Ommer, {Taming transformers for high-resolution image synthesis}, in: Proceedings of the IEEE/CVF conference on computer vision and pattern recognition, 2021, pp. 12873--12883.

\bibitem{sber-vqgan}
github.com, \href{https://github.com/ai-forever/MoVQGAN}{{SBER-MoVQGAN}} (2024).
\newline\urlprefix\url{https://github.com/ai-forever/MoVQGAN}

\bibitem{kandinsky-2-prior-weights}
huggingface.co, \href{https://huggingface.co/kandinsky-community/kandinsky-2-2-prior}{{Hugging Face Hub Kandinsky Community Kandinsky 2.2 Prior}} (2024).
\newline\urlprefix\url{https://huggingface.co/kandinsky-community/kandinsky-2-2-prior}

\bibitem{kandinsky-2-decoder-weights}
huggingface.co, \href{https://huggingface.co/kandinsky-community/kandinsky-2-2-decoder}{{Hugging Face Hub Kandinsky Community Kandinsky 2.2 Decoder}} (2024).
\newline\urlprefix\url{https://huggingface.co/kandinsky-community/kandinsky-2-2-decoder}

\bibitem{hugging-face}
huggingface.co, \href{https://huggingface.co/docs/hub/index}{{Hugging Face Hub documentation}} (2024).
\newline\urlprefix\url{https://huggingface.co/docs/hub/index}

\bibitem{flux-weights}
huggingface.co, \href{https://huggingface.co/black-forest-labs/FLUX.1-dev}{{Hugging Face Hub FLUX.1-dev}} (2024).
\newline\urlprefix\url{https://huggingface.co/black-forest-labs/FLUX.1-dev}

\bibitem{medfusion}
G.~M{\"u}ller-Franzes, J.~M. Niehues, F.~Khader, S.~T. Arasteh, C.~Haarburger, C.~Kuhl, T.~Wang, T.~Han, T.~Nolte, S.~Nebelung, J.~N. Kather, D.~Truhn, {A multimodal comparison of latent denoising diffusion probabilistic models and generative adversarial networks for medical image synthesis}, Sci. Rep. 13~(1) (2023) 12098.

\bibitem{vae}
D.~P. Kingma, M.~Welling, {Auto-encoding variational bayes}, arXiv preprint arXiv:1312.6114.

\bibitem{clip-site}
openai.com, \href{https://openai.com/index/clip/}{{CLIP: Connecting text and images}} (2024).
\newline\urlprefix\url{https://openai.com/index/clip/}

\bibitem{zhang2024biomedclip}
S.~Zhang, Y.~Xu, N.~Usuyama, H.~Xu, J.~Bagga, R.~Tinn, S.~Preston, R.~Rao, M.~Wei, N.~Valluri, C.~Wong, A.~Tupini, Y.~Wang, M.~Mazzola, S.~Shukla, L.~Liden, J.~Gao, A.~Crabtree, B.~Piening, C.~Bifulco, M.~P. Lungren, T.~Naumann, S.~Wang, H.~Poon, \href{https://ai.nejm.org/doi/full/10.1056/AIoa2400640}{{A Multimodal Biomedical Foundation Model Trained from Fifteen Million Image–Text Pairs}}, NEJM AI 2~(1).
\newblock \href {http://dx.doi.org/10.1056/AIoa2400640} {\path{doi:10.1056/AIoa2400640}}.
\newline\urlprefix\url{https://ai.nejm.org/doi/full/10.1056/AIoa2400640}

\bibitem{adamw}
I.~Loshchilov, F.~Hutter, {Decoupled weight decay regularization}, arXiv preprint arXiv:1711.05101.

\bibitem{cluster-link}
P.~S. Kostenetskiy, R.~A. Chulkevich, V.~I. Kozyrev, \href{https://dx.doi.org/10.1088/1742-6596/1740/1/012050}{{HPC Resources of the Higher School of Economics}}, Journal of Physics: Conference Series 1740~(1) (2021) 012050.
\newblock \href {http://dx.doi.org/10.1088/1742-6596/1740/1/012050} {\path{doi:10.1088/1742-6596/1740/1/012050}}.
\newline\urlprefix\url{https://dx.doi.org/10.1088/1742-6596/1740/1/012050}

\end{thebibliography}

\end{document}